\documentclass[12pt]{amsart}

\usepackage{graphicx}
\usepackage{grffile}

\usepackage{epstopdf}
\DeclareGraphicsExtensions{.png,.pdf}

\usepackage[utf8]{inputenc}
\usepackage[T1]{fontenc}

\usepackage{hyperref}
\usepackage{amsmath}
\usepackage{amscd}
\usepackage{pb-diagram}
\usepackage{comment}

\usepackage{subfig}
\usepackage{amssymb}
\usepackage{lmodern}
\usepackage{color}
\usepackage{wrapfig}
\def\corcommstyle#1{\bf\small\tt\textcolor{blue}{#1}}

\def\corrl #1<<#2||#3>>{
\if\visiblecomments y
  \begin{quote} {\corcommstyle{ $<<$COMMENT$>>$ #1\marginpar{!!}\\$<<$OLD$<<$}} \end{quote}
  #2
  \begin{quote} {\corcommstyle{ ==NEW== }} \end{quote}
  #3
  \begin{quote} {\corcommstyle{ $>>$END$>>$ }} \end{quote}
 \else
  #3
 \fi
}

\long\def\longcorrl #1<<#2||#3>>{
\if\visiblecomments y
  \begin{quote} {\corcommstyle{ $<<$COMMENT$>>$ #1\marginpar{!!}\\$<<$OLD$<<$}} \end{quote}
  #2
  \begin{quote} {\corcommstyle{ ==NEW== }} \end{quote}
  #3
  \begin{quote} {\corcommstyle{ $>>$END$>>$ }} \end{quote}
 \else
  #3
 \fi
}

\def\corrq #1<<#2>>{
\if\visiblecomments y
  \begin{quote} {\corcommstyle{ $<<$COMMENT$>>$ #1\marginpar{!!}\\$<<$BEG$<<$}} \end{quote}
  #2
  \begin{quote} {\corcommstyle{ $>>$END$>>$ }} \end{quote}
 \else
  #2
 \fi
}

\long\def\longcorrq #1<<#2>>{
\if\visiblecomments y
  \begin{quote} {\corcommstyle{ $<<$COMMENT$>>$ #1\marginpar{!!}\\$<<$BEG$<<$}} \end{quote}
  #2
  \begin{quote} {\corcommstyle{ $>>$END$>>$ }} \end{quote}
 \else
  #2
 \fi
}

\def\corrc #1<<>>{
\if\visiblecomments y
  \begin{quote} {\corcommstyle{ $<<$COMMENT$>>$ #1\marginpar{!!}}} \end{quote}
\fi
}

\def\corrse #1<<>>{
\if\visiblecomments y
  \begin{quote} {\corcommstyle{ $<<$SECOND EDITION$>>$ #1\marginpar{!!}}} \end{quote}
\fi
}

\def\corre #1<<#2||#3>>{
\if\visiblecomments y
  #3\marginpar{\corcommstyle{ #1}}
 \else
  #3
 \fi
}

\long\def\longcorre #1<<#2||#3>>{
\if\visiblecomments y
  #3\marginpar{\corcommstyle{ #1}}
 \else
  #3
 \fi
}

\def\corrs #1<<#2||#3>>{
\if\visiblecomments y
  \textcolor{red}{#3}\marginpar{\corcommstyle{ #2 $\rightarrow$ #3\\ #1}}
 \else
  #3
 \fi
}

\def\corro #1<<#2||#3>>{
#2}

\def\corrn #1<<#2||#3>>{
#3}

\long\def\longcorro #1<<#2||#3>>{
#2}

\long\def\longcorrn #1<<#2||#3>>{
#3}

\long\def\underconstruction #1<<<#2>>>{
\if\visiblecomments y
  \begin{quote} {\corcommstyle{ $<<$UNDER CONSTRUCTION - BEGIN$>>$ #1\marginpar{!!}}} \end{quote}
  #2
  \begin{quote} {\corcommstyle{ $>>$UNDER CONSTRUCTION - END$>>$ }} \end{quote}
 \else
 \fi
}

\def\showcomments{
  \let\visiblecomments y
}

\def\hidecomments{
  \let\visiblecomments n
}

\let\visiblecomments y

\def\refeq#1{\if\workingver y(\ref{#1})-[[#1]]\else(\ref{#1})\fi}
\def\refth#1{\if\workingver y\ref{#1}-[[#1]]\else\ref{#1}\fi}
\def\mylabel#1{\if\workingver y\label{#1}{\bf\ \ [[#1]]\ \ }\else\label{#1}\fi}
\def\mybibitem#1{\if\workingver y\bibitem{#1}{\bf\ \ [[#1]]\ \
}\else\bibitem{#1}\fi}

\newfont{\msam}{msam10}
\newfont{\msbm}{msbm10}

\def\cC{\text{$\mathcal C$}}

\def\cG{\text{$\mathcal G$}}

\def\cL{\text{$\mathcal L$}}

\def\cR{\text{$\mathcal R$}}

\def\cT{\text{$\mathcal T$}}

\def\cV{\text{$\mathcal V$}}
\def\cW{\text{$\mathcal W$}}

\def\bP{\text{$\mathbf P$}}
\def\bQ{\text{$\mathbf Q$}}
\def\bR{\text{$\mathbf R$}}
\def\bS{\text{$\mathbf S$}}

\newcommand{\cl}{\operatorname{cl}}

\renewcommand{\emptyset}{\varnothing}
\renewcommand{\subset}{\subseteq}

\def\begeq#1{\begin{equation}\mylabel{#1}}
\def\endeq{\end{equation}}

\def\mathobj#1{\mbox{$#1$}}

\def\NN{\mathobj{\mathbb{N}}}

\def\RR{\mathobj{\mathbb{R}}}

\newcommand{\mto}{\multimap}

\usepackage[all]{xy}
\usepackage{algorithm}
\usepackage[noend]{algpseudocode}
\usepackage{caption}
\usepackage{float}
\usepackage{boldline}

\usepackage[T1]{fontenc}
\usepackage{yfonts}

\usepackage{mathbbol} 
\usepackage{yfonts}

\usepackage[thinc]{esdiff}

\usepackage{adjustbox}
\def\gW{\text{$\textgoth W$}}
\def\gV{\text{$\textgoth V$}}

\newcommand{\corpus}[5]{{ \cL^{#1,#2}\gW^{#3,#4}_{#5}  }}

\definecolor{darkgreen}{rgb}{0.0, 0.8, 0.0}
\definecolor{darkred}{rgb}{1, 0.1, 0.3}
\definecolor{darkblue}{rgb}{0.1, 0.1, 1}

\newcommand{\cmf}{\texttt{cmf}}
\newcommand{\cmfgraph}{\texttt{cmf graph}}
\newcommand{\sorted}{\operatorname{sorted}}
\newcommand{\hash}{\operatorname{\textgoth{h}}} 
\newcommand{\Rgrid}{\operatorname{\cR}} 

\title{Unsupervised Features Learning for Sampled Vector Fields}

\author[M.\ Juda]{Mateusz Juda}
\address{Mateusz Juda, Division of Computational Mathematics,
  Institute of Computer Science and Computational Mathematics,
  Faculty of Mathematics and Computer Science,
  Jagiellonian University, ul.~St. \L{}ojasiewicza 6, 30-348~Kra\-k\'ow, Poland.
}
\email{mateusz.juda@ii.uj.edu.pl}
\thanks{The author is supported
       by the Polish National Science Center under Ma\-estro Grant 2014/14/A/ST1/00453.}

\date{}

\begin{document}

\begin{abstract}
In this paper we introduce a new approach to computing hidden features of sampled vector fields. The basic idea is to convert the vector field data to a graph structure and use tools designed for automatic, unsupervised analysis of graphs. Using a few data sets we show that the collected features of the vector fields are correlated with the dynamics known for analytic models which generate the data. In particular the method may be useful in analysis of data sets where the analytic model is poorly understood or not known.
\end{abstract}

\maketitle

\section{Introduction}
\label{sec:intro}
Continuous mathematical models are useful to analyze and draw conclusions about complicated physical systems, where values of system states are assumed to be real numbers.
However, nowadays  we have countless possibilities of data collection, so scientific and industrial challenges are mostly data driven. We have only finite amount of information, so models should be well-fitted to the observed data and also adapt properly to new, previously unseen data. Usually it means that we have to create highly parameterized models in an automatic way.

We propose a new method for automatic modeling of vector field data sets. In particular, our method takes as an input a finite collection of vectors and creates a low dimensional description of the data. The description encodes features of the observed vector field and allows us to further analyze the physical system using a smaller amount of information.
We only require a point cloud with vectors attached at each point. There is no need to manually model the system, e.g. via differential equations.
Thanks to this we can analyze data generated by processes for which it is hard to create a traditional model, for instance magnetic field on the Sun surface~\cite{scherrer_helioseismic_2012}.

The goal of this paper is to present the new method and to validate it on well understood data sets. We show examples based mostly, but not only, on simulated dynamical systems. We compare the automatically learned features with well known, analytically calculated, properties of the systems. We also extend the analysis to a series of dynamical systems, either given as parameterized equations or vector fields constructed from data. The presented examples justify the usefulness of the methods. The learned features may be used as an input to other machine learning tasks, where the original data may be viewed as vector fields, e.g. solar flares predictions, classification of data from particle image velocimetry (PIV), turbulences detection. Details of the applications are beyond the scope of this paper, are in progress, and are going to be presented elsewhere.

The paper is organized as follows: Section~\ref{sec:intro} contains the introduction, a description of related work, and it recalls the theoretical model. In Section~\ref{sec:desc} we describe the problem and model it as the problem of word embedding in Natural Language Processing (NLP). In Section~\ref{sec:alg} we describe the details of the algorithm. Finally, Section~\ref{sec:ex} contains examples and Section~\ref{sec:conc} finalizes the paper.

\subsection{Related work}
The computation of global dynamical information is a challenging problem for applications. Characterization of the global dynamical structure and its changes are fundamental in many disciplines, e.g. computational biology and engineering. Combinatorial dynamics and computational topology are powerful tools for the task. In particular, the tools are useful in classification of the qualitative properties of parameterized models. The database approach~\cite{arai_database_2009} helps to understand models where it is difficult to measure parameters. For sampled parameters an outer approximation of the dynamics is computed using rigorous numerical methods. The dynamics is then represented as a directed graph and classified using the Morse decomposition and Conley indices.
The method is useful when we know the parameterized dynamical system model. Then, by rigorous simulations, we can find all possible dynamics and match them with collected data.
Recent application of the method is applied to understand the global dynamics of gene regulatory networks~\cite{cummins_combinatorial_2016, gedeon_identifying_2018}.

In experiments very often one quantity is measured - a time series - while in numerical simulations the full state of a system is an observable. A powerful tool to reason about the unknown system using only a partial information is the Takens embedding theorem~\cite{takens_detecting_1981}. The theorem has been used in~\cite{batko_conley_2019, mischaikow_construction_1999}, together with the Conley index theory of multivalued maps, to identify dynamics from sampled data.

A different approach to sampled dynamical systems is presented in~\cite{edelsbrunner_persistent_2015}. The method analyze data points given by an unknown self-map. The results presented in this paper suggest that the persistent homology of eigenspaces picks up the important dynamics from small data sample.

Problems similar to our work are considered in~\cite{szymczak_robust_2012} where the input vector field is converted to a piecewise constant vector field. Then the Morse decomposition is computed for trajectories obtained using geometrical rules. A main difference, when compared to our work, is that the method is limited to triangulated manifold surfaces and considers only the Morse decomposition while we show a more general framework.

\subsection{A combinatorial dynamical system from a sampled vector field}
In this section we recall some definitions and results from~\cite{dey_persistent_2019,mrozek_conleymorseforman_2017}. As mentioned in~\cite{dey_persistent_2019} also here the presented results may be generalized to arbitrary finite $T_0$ topological spaces~\cite{barmak_algebraic_2011}.
From the viewpoint of applications, a finite topological space may be a collection of cells of a simplicial, cubical, or general cellular complex approximating a cloud of sampled points. For the sake of this paper we use simplicial complexes only.

Let $K$ be a finite simplicial complex, either a geometric simplicial complex in $\RR^d$ or an abstract simplicial complex (see \cite[Section 1.2, 1.3]{munkres_elements_2018}).
We consider $K$ as a poset $(K,\preceq)$ with $\sigma\preceq\tau$ if and only if $\sigma$ is a face of $\tau$
(also phrased $\tau$ is a {\em coface} of $\sigma$).
The poset structure of $K$ provides, via the Alexandrov Theorem~\cite{alexandroff_diskrete_1937}, a $T_0$ topology on $K$.
We say that $A\subset  K$ is {\em orderly convex} if for any $\sigma_1,\sigma_2 \in A$ and $\tau\in  K$ the relations $\sigma_1\preceq \tau$ and $\tau\preceq \sigma_2$ imply $\tau\in A$.
We remark that orderly convex sets in $K$ may be characterized in the language of  the associated Alexandrov topology.
Namely, $A\subset K$ is orderly convex if and only if it is locally closed (see \cite[Sec. 2.7.1, p. 112]{engelking_general_1989}) in the Alexandrov topology $\cT_K$.

We define a {\em multivector} as an orderly convex subset of $K$ and
a {\em combinatorial multivector field} (\cmf{} in short) on $K$ as a partition $\cV$ of $K$ into multivectors.
A multivector is {\em critical} if its Lefschetz homology is non zero (for details see~\cite{mrozek_conleymorseforman_2017}).
The definition encompass the combinatorial vector field of Forman~\cite{forman_combinatorial_1998,forman_morse_1998} as a special case.

Given a \cmf{} $\cV$, we denote by $[\sigma]_\cV$ the unique $V$ in $\cV$ such that $\sigma\in V$. We associate with $\cV$ a {\em combinatorial dynamical system} $F_\cV: K\mto  K$ (a multi-map) given by
$
   F_\cV(\sigma):=\cl \sigma \cup [\sigma]_\cV,
$
   where $\cl \sigma$ is the {\em clousure} of $\sigma$ defined by $\cl \sigma  := \{ \tau \mid \tau \preceq \sigma \}$.

When the dynamics which is sampled constitutes of a flow, that is, when time is continuous as in the case of a differential equation, the sampled data
may consist of a cloud of points with a vector attached to every point.
We call this a {\it cloud of vectors}.
In this case the construction of a combinatorial dynamical system is done in two
steps. In the first step, the cloud of vectors is transformed into a combinatorial vector field
in the sense of Forman~\cite{forman_combinatorial_1998,forman_morse_1998} or its generalized version of a combinatorial multivector field~\cite{mrozek_conleymorseforman_2017}.
In the second step, the combinatorial multivector field is transformed into a combinatorial dynamical system.
Intuitively, cells of $K$ fill the gaps between the points from the input data. The multivectors reflect the vector field behavior from lower to higher dimensional cells. The value of $F_\cV(\sigma)$ allows us to travel between the points. We can use either the fillings $[\sigma]_\cV$ or jump into an area influenced by another vector using $\cl \sigma$.

\begin{figure*}[!htbp]
  \begin{center}
    \raisebox{0.02\height}{\includegraphics[width=0.23\textwidth]{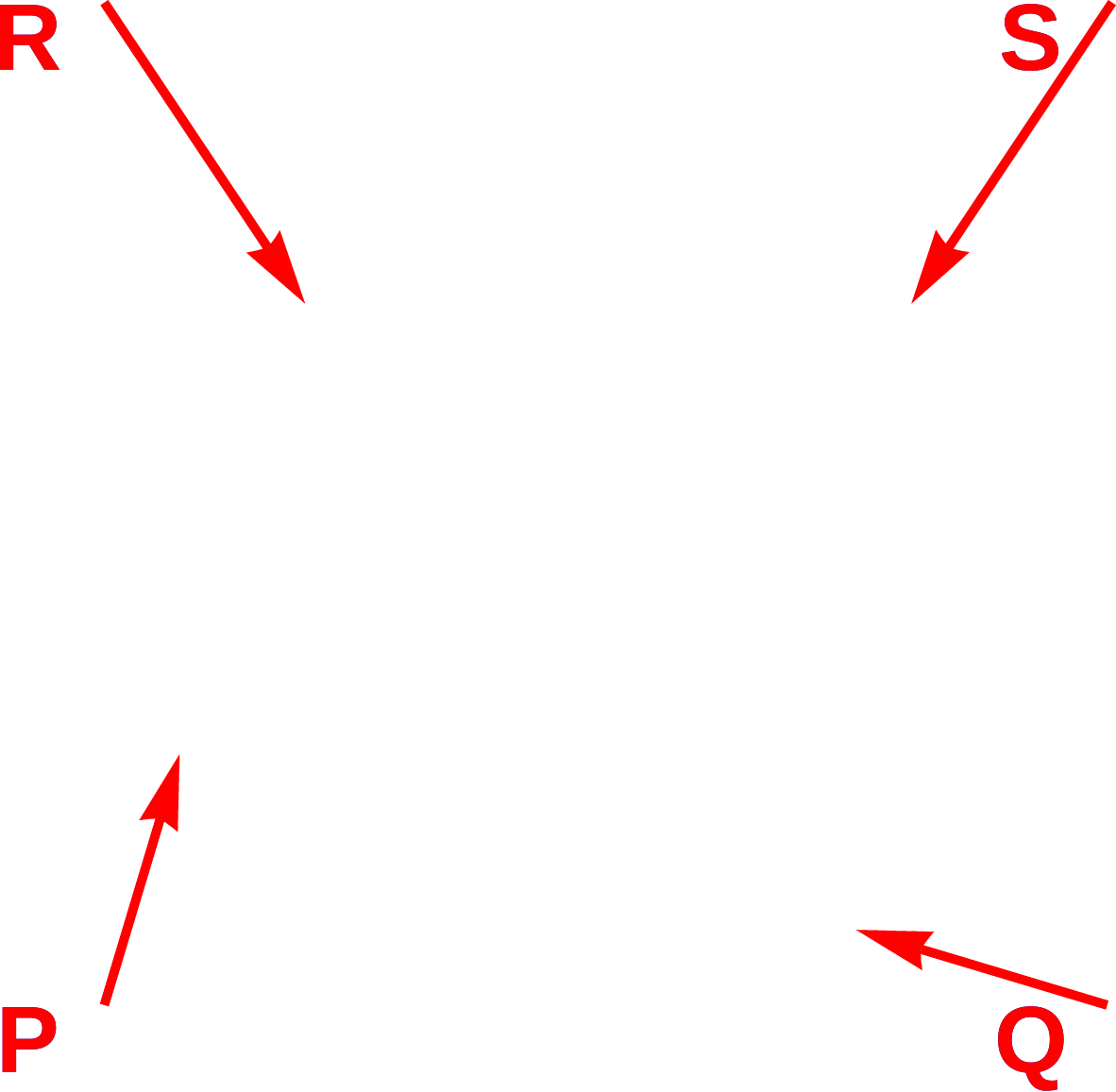}}\hspace*{15pt}
    \raisebox{0.02\height}{\includegraphics[width=0.23\textwidth]{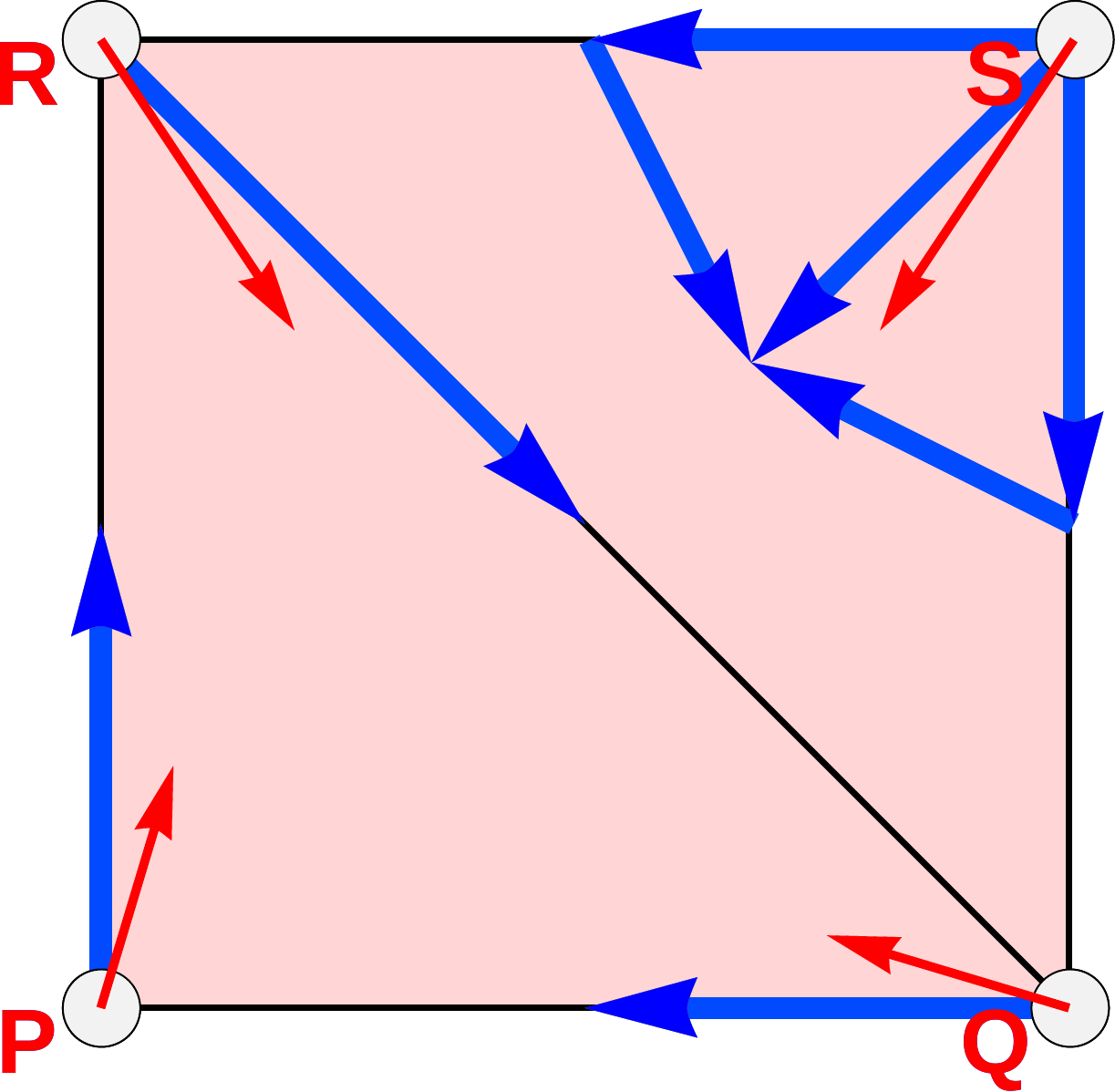}}\hspace*{10pt}
    \includegraphics[width=0.4\textwidth]{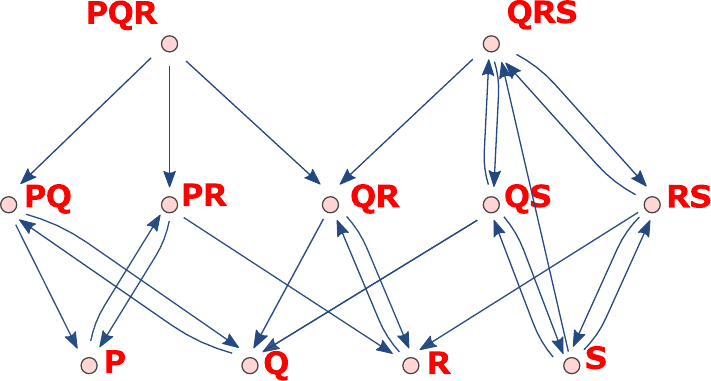}
  \end{center}
  \caption{Left: A cloud of vectors.
  Middle: A possible combinatorial multivector field representation of the cloud of vectors.
    Right: The associated combinatorial dynamical system represented as a digraph.}
  \label{fig:fv-map}
\end{figure*}

Figure~\ref{fig:fv-map}(left) recall a toy example of a cloud of vectors~\cite{dey_persistent_2019}.
It consists of four vectors marked red at four points $\bP$, $\bQ$, $\bR$, $\bS$.
One of possible geometric simplicial complexes with vertices
at points $\bP$, $\bQ$, $\bR$, $\bS$ is the simplicial complex $K$ consisting of
triangles $\bP\bQ\bR$, $\bQ\bR\bS$ and its faces.
A possible multivector field $\cV$ on $ K$ constructed from the cloud of vectors consists of multivectors $\{\bP,\bP\bR\}$, $\{\bR,\bQ\bR\}$, $\{\bQ,\bP\bQ\}$, $\{\bP\bQ\bR\}$,$\{\bS,\bR\bS,\bQ\bS,\bQ\bR\bS\}$.
This multivector field is indicated in Figure~\ref{fig:fv-map}(middle) by blue
arrows between centers of mass of simplices.
Note that in order to keep the figure legible, only arrows in the direction
increasing the dimension are marked.
The associated combinatorial dynamical system $F_\cV$ presented as a digraph is in Figure~\ref{fig:fv-map}(right).
Note that in general  $K$ and $\cV$ are  not uniquely determined by the cloud of vectors.
One possible method for constructing combinatorial multivector fields from a cloud of vectors is discussed in ~\cite[Section 7.2]{dey_persistent_2019}.

Given a combinatorial multivector field $\cV$, we define a {\em combinatorial multivector field graph} (\cmfgraph{} in short), denoted by $G_{\cV}$, as a directed graph with the set of vertices $V(G_{\cV}) := \cV$ and the set of  edges $E(G_{\cV}) := \{([u]_\cV, [v]_\cV) \mid v\in F_{\cV}(u)\})\}$. For a graph $G$ by $G^T$ we denote the {\em transpose} of $G$, i.e.  $G^T$ is a graph, such that $V(G^T) := V(G)$ and $E(G^T) := \{ ([v]_\cV, [u]_\cV) \mid ([u]_\cV,[v]_\cV) \in E(G)\}$.
We define the {\em forward distance} (resp. {\em backward distance}) between two vertices as the number of edges in a shortest path connecting them in $G_{\cV}$ (resp. $G_{\cV}^T$).

\section{Problem description}
\label{sec:desc}
Our goal is to explore combinatorial multivector field graphs structure. We do it, by finding features of multivectors, as fixed length sequences of real numbers, such that multivectors with a similar local structure in a \cmfgraph{}  have similar features. We also want to extend the features to whole graphs, in such a way that \cmfgraph{}s representing similar combinatorial dynamical systems have similar features.
At this point we skip formal definitions of the similarities mentioned above and we use experimental justifications of the presented methods. A theory of a similarity for combinatorial dynamical systems is not developed yet and the topic is still under research. However, as long as a combinatorial dynamical system $F_\cV$ reflects the dynamics of the sampled system, tools designed for graphs can be extended to study combinatorial dynamical systems.

Let $\cG$ be a collection of labeled graphs, namely each vertex $v$ of a graph $G\in \cG$ has a label $l_v$ from some set of labels $L \subset \NN$. In Section~\ref{sec:voc} we propose an assignment of labels which uses topological properties of the multivectors.
Our main goal is to {\it learn} latent representations of labels of a \cmfgraph{} $G_{\cV}$. Namely, we want to find an {\em encoding function} $\Phi: L \to \RR^D$, for a fixed dimension $D$, such that the codes $\Phi(l_u)$ and $\Phi(l_v)$ are close whenever there is a similar local structure in  $G_{\cV}$ around two multivectors $u$ and $v$.
We want to extend the encoding function to graphs, namely for two \cmf{} $\cV_1$ and $\cV_2$ the codes of $\Phi(G_{\cV_1})$ and $\Phi(G_{\cV_2})$ are close whenever the dynamical systems which generates $\cV_1$ and $\cV_2$ are similar. The values of $\Phi$ should depend on qualitative local features of the vertices and do not depend on the vertices enumeration.

Let $W_v = \{w_1, w_2, \ldots, w_k\} \subset G$ be a {\em random walk} rooted at $v \in G \in \cG$, i.e. $v = w_1$ and $w_{i+1}$ is a randomly selected neighbor of $w_i$. We explore the graph $\cG$ using the {\em DeepWalk} technique, namely short random walks, as described in~\cite{perozzi_deepwalk:_2014}.
The idea behind the approach is to generalize Natural Language Processing (NLP) methods to explore graphs. In this setting we treat labels of vertices as words  and the random walks as sentences in an artificial language. Afterwards, we use the {\em Continuous Skip-gram Model}~\cite{mikolov_efficient_2013} to analyze the text documents structure and to find $\Phi$, such that it minimize the following log probability:
\begin{equation}
  \label{eq:prob_max}
  \underset{\Phi}{\text{minimize}} - \log P( \{l_{w_{i-w}}, \ldots, l_{w_{i+w}}\} \setminus \{ l_{w_i} \} \mid \Phi(l_{w_i}) ),
\end{equation}
where $w$ is a parameter called the {\em window size}.
As we can see in \eqref{eq:prob_max} the goal is to predict context based on a word without taking into account the order of words.
A relaxation scheme described in~\cite{mikolov_efficient_2013,mikolov_distributed_2013} provides efficient algorithms to compute the latent representation of words, the encoding $\Phi$.

\section{Algorithm}
\label{sec:alg}

In the context of this paper we assume a family $\gV$ of combinatorial multivector fields is given. Having a \cmf{} $\cV \in \gV$ we transform it to the NLP data using the following steps:
\begin{enumerate}
\item assign labels (words) to multivectors,
\item using short random walks encode the structure of $\cV$ as a text document,
\item using NLP techniques analyze the document and extract information about the multivector fields.
\end{enumerate}
Before we present detailed description of the steps we want to recall a few programming tools.

For a set $A \subset \NN$ by $\sorted(A)$ we mean a sequence $\{a_1, a_2,\ldots,a_n\}$ such that $a_i$ in $A$ and $a_i \le a_j$ for each $i < j$. We recall that a {\it hash function} is any function $\hash$ that projects data of an arbitrary size to a value from a set with a fixed number of members~\cite{cormen_introduction_2009, knuth_art_1998}. A {\it good} hash function satisfies the following properties: 1) it is fast to compute; 2) it minimize {\it collisions}, i.e. duplication of the function values.
In practice programming languages (or additional libraries) implement a hash function for each built-in data type.
For a user-defined data type a hash function may be easily defined using hashes of the data type members, e.g. for a pair of numbers $(a,b)$ we may define $\hash ( (a,b) ) := \hash(a) \text{{\bf xor}} \hash(b)$, where  $\text{{\bf xor}}$ is the bitwise exclusive OR operation. To simplify the notation we use the symbol $\hash$ regardless of the function domain. We assume that a good hash function $\hash$ with $64$-bits values is given for the following data types:
natural numbers, tuples of natural numbers, list of natural numbers, and lists of lists of natural numbers.

\subsection{Topological vocabulary}
\label{sec:voc}
The NLP procedure requires a vocabulary in order to assign labels to the vertices of a graph.
We construct labels which grasp some local, topological properties of the vertices in the vector field.
Intuitively, we obtain a labeling which is universal, i.e. can be computed using only a multivector and its neighbors and does not depend on the global dynamics. It is only an example of many possible labelings.

For a multivector $v \in \cV$, that is a vertex in $G_\cV$, we first define the {\em label of $v$ at level (0,0)}, denoted by $\cL^{0,0}(v)$, in the following way:
\[
  \cL^{0,0}(v) := \hash (\max_{\sigma \in v} \dim \sigma, \vert v \vert, \chi(v)),
\]
where $\dim \sigma$ denotes the dimension of a cell $\sigma$, $\vert v \vert$ stands for the cardinality of $v$, and $\chi(v)$ is the Euler characteristic of $v$ (i.e. $\sum_{\sigma \in v} (-1)^{\dim \sigma}$).
We use the $\hash$ function in the above definition to obtain a number as the label of $v$. Formally it is not required, however it simplified the computations and the notation.

We define the {\em label of $v$ at level (b,f)} , denoted by $\cL^{b,f}(v)$, in the following way:
\[
\cL^{b,f}(v) := \hash (\cL^{0,0}(v), \sorted(\{\cL^{0,0}(u) \mid u \in N^{+}_{f}(v)\}), \sorted(\{\cL^{0,0}(u) \mid u \in N^{-}_{b}(v)\})),
\]
where $N^{+}_{f}(v)$ (resp. $N^{-}_{b}(v)$) are sets of vertices in the forward (resp. backward) distance from $v$ not bigger than $f$ (resp. $b$).
We use the $\sorted$ function to unify the order of the neighbors in the sets $N^{+}_{f}(v)$ and $N^{-}_{b}(v)$.
We emphasize that the values of $\cL^{0,0}$ are in $\NN$ because of the hash function, so the input for the $\sorted$ function is a set of numbers. Also the value of $\cL^{b,f}$ is always a number because of the hash function.

Currently we use NLP methods which do not use the words (labels) structure. It means that we can arbitrary map the tuples to numbers using a reasonable good hashing function $\hash$. However, there are NLP methods which operates on sub-words ($n$-grams)~\cite{bojanowski_enriching_2017} and more sophisticated labelings may take advantage of a multivector neighborhood structure.

\begin{figure*}[!htbp]
  \begin{center}
    \includegraphics[width=0.2\textwidth]{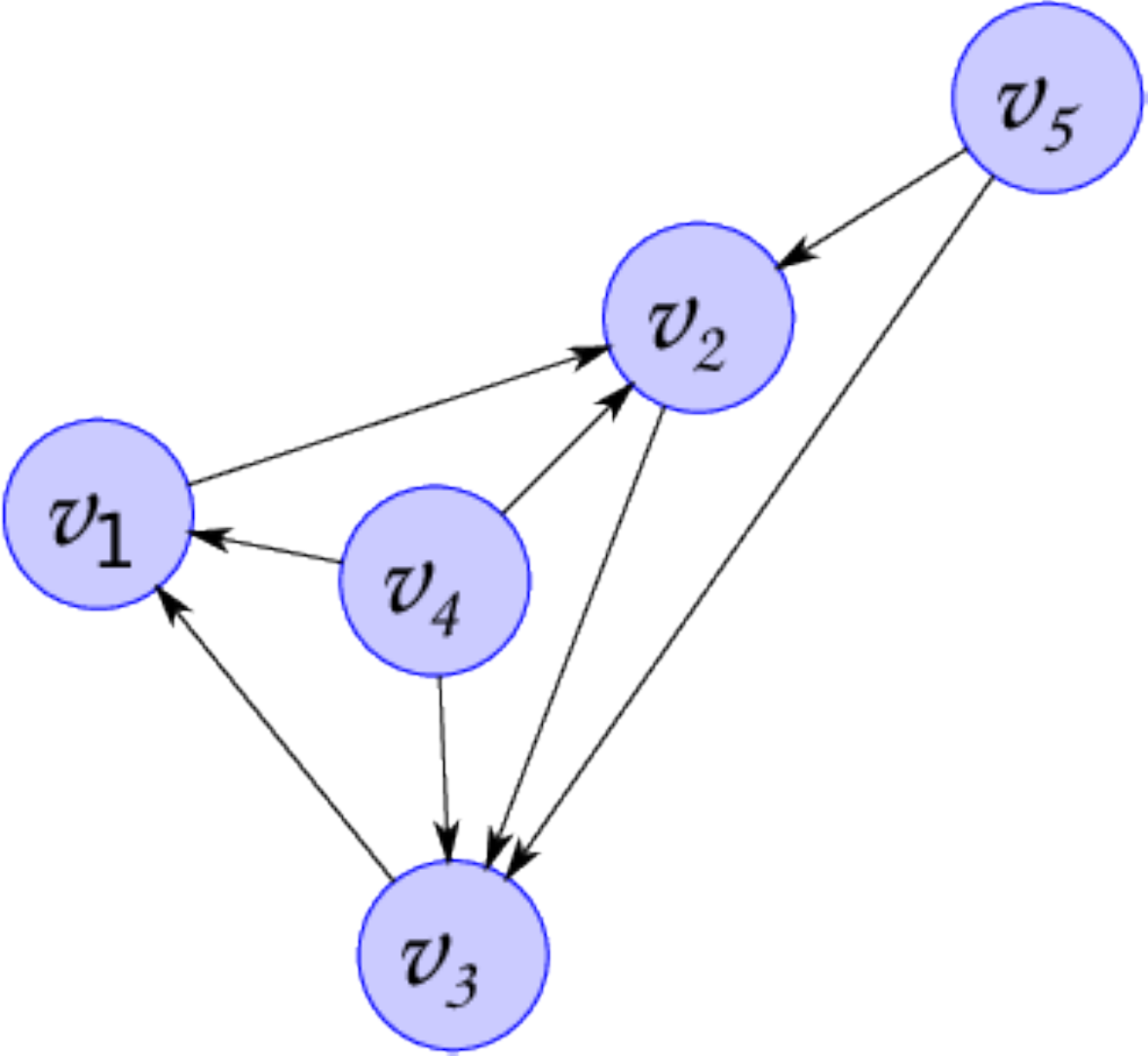}
  \end{center}
  \caption{$G_\cV$ graph for the example presented in Figure~\ref{fig:fv-map}. The nodes represent multivectors in the following order:
    $v_1 = \{\bP,\bP\bR\}$,
    $v_2 = \{\bR,\bQ\bR\}$,
    $v_3 = \{\bQ,\bP\bQ\}$,
    $v_4 = \{\bP\bQ\bR\}$,
    $v_5 = \{\bS,\bR\bS,\bQ\bS,\bQ\bR\bS\}$.
  }
  \label{fig:mv-graph}
\end{figure*}

As an example we consider the multivector field  $\cV$  and its combinatorial dynamical system $F_{\cV}$ presented in Figure~\ref{fig:fv-map}. Figure~\ref{fig:mv-graph} presents the associated graph on multivectors $G_\cV$. Table~\ref{table:nlp} presents step by step calculations of the values of $\cL^{1,1}$.

\begin{table}[H]
\small
\centering
\begin{adjustbox}{width=\textwidth}
\begin{tabular}{|c|c|c|c|c|l|}
  \hline
  $v$ & simplices of $v$ & $\cL^{0,0}(v)$ & $N^{+}_{1}(v)$ & $N^{-}_{1}(v)$ & $\cL^{1,1}(v)$ \\
  \hline \hline
  $v_1$ & $\{ \bP, \bP\bR \}$ & $\hash (1,2,0) = 5095$ & $\{v_2\}$ & $\{v_3, v_4\}$ &

$\begin{array}{l l}
   \hash ( &  \\
     & \hash(1, 2, 0),\\
     & \sorted( \hash(1,2,0)), \\
   & \sorted(\hash(1,2,0), \hash(2,1,1)) \\
   ) = & 4901
\end{array}$
  \\
  \hline
  $v_2$ & $\{ \bR, \bQ\bR \}$ & $\hash (1,2,0)  = 5095$ & $\{v_3\}$ & $\{v_1, v_4, v_5\}$ &
  $\begin{array}{l l}
    \hash ( & \\
    & \hash(1, 2, 0), \\
    & \sorted(\hash(1,2,0)), \\
    & \sorted (\hash(1,2,0), \hash(2,1,1), \hash(2,4,0)) \\
    ) = & 7355
  \end{array}$
  \\
  \hline
  $v_3$ & $\{ \bQ, \bQ\bP \}$ & $\hash (1,2,0)  = 5095$ & $\{v_1\}$ & $\{v_2, v_4, v_5\}$ &
    $\begin{array}{l l}
    \hash ( & \\
    & \hash(1, 2, 0),\\
    & \sorted(\hash(1,2,0)), \\
    & \sorted( \hash(1,2,0), \hash(2, 1, 1), \hash(2,4,0)) \\
    ) = & 7355
  \end{array}$
  \\
  \hline
  $v_4$ & $\{ \bP\bQ\bR \}$ & $\hash (2,1,1) = 6161$ & $\{v_1, v_2, v_3\}$ & $\emptyset$ &
  $\begin{array}{l l}
    \hash ( & \\
    & \hash(2, 1, 1), \\
    & \sorted(\hash(1,2,0), \hash(1,2,0), \hash(1,2,0)), \\
    & \emptyset \\
    ) = & 5836
  \end{array}$
  \\
  \hline
  $v_5$ & $\{\bS,\bR\bS,\bQ\bS,\bQ\bR\bS\}$ & $\hash (2,4,0) = 6275$ & $\{v_2, v_3\}$ & $\emptyset$ &
  $\begin{array}{l l}
    \hash ( & \\
    & \hash(2, 4, 0), \\
    & \sorted(\hash(1,2,0), \hash(1, 2, 0)),\\
    & \emptyset \\
    ) = & 5382
  \end{array}$
  \\
  \hline
\end{tabular}
\end{adjustbox}
\caption{Step by step calculation of the labels at level $1$ for the example presented in Figure~\ref{fig:fv-map} and Figure~\ref{fig:mv-graph}. We use the standard implementation of $\texttt{hash}$ and $\texttt{sorted}$ functions from the Python programming language (for simplicity the values are taken modulo $10^4$).
  We notice that the labels at level $(0,0)$  cannot distinguish multivectors $v_1, v_2, v_2$. However, for the labels at level $(1,1)$ the labeling of $v_1$ is different from the labelings of $v_2$ and $v_3$. }
\label{table:nlp}
\end{table}

\subsection{Corpus}
\label{sec:corpus}
A {\em $d$-random multivector walk} {\em on} $G_{\cV}$ {\em from} $s$,
denoted by $\cW_{\cV}^{d+}(s)$,
is a stochastic process with random variables $\{ W_1, W_2, \ldots, W_d\}$ such that $W_1 = s$, and $W_{i+1}$ is a vertex chosen at random from the set $N^{+}_{1}(W_{i}) \cup \{W_{i}\}$. The probability $P(W_{i+1} = u)$ is defined as $p_u/\sum_{v}{p_v}$, where
  \begin{equation}
    p_v=
    \begin{cases}
      1, & \text{if } v \in N^{+}_{1}(W_{i}) \text{ and } v \ne W_i\\
      1, & \text{if } v = W_i \text{ and } W_i \text{ is critical}\\
      0, & \text{otherwise.}
    \end{cases}
  \end{equation}
In the above definition we can replace $N^+$ with $N^-$ and reverse the order of the random multivector walk. This way we define a {\em $d$-random multivector walk} {\em on} $G_{\cV}$ {\em to} $t$, denoted by $\cW_{\cV}^{d-}(t)$.

Let $\gV$ be a collection of combinatorial multivector fields.
A {\em $(c,d)$-random multivector corpus} of $\gV$
is the stochastic process:
\[
\gW^{c,d}_\gV := \bigoplus_{\substack{ \cV \in \gV \\ u \in \cV\\ \bullet \in \{+,-\} \\ i \in {1,\ldots,c} } } \cW_{\cV}^{d\bullet}(u),
\]
where $\bigoplus$ denotes concatenation of sequences of the random variables.
Note that in the above definition we repeat $c$ times the random multivector walk.
We define a {\em multivector $(f, b, c, d)$-corpus} of $\gV$, denoted by  $\corpus{f}{b}{c}{d}{\gV}$, as
$\{ \cL^{f,b}(W) \mid W \in \gW^{c,d}_\gV  \}$, where $\cL^{f,b}(W)$ is a label of $W$ at level $(f, b)$ (as defined in Section~\ref{sec:voc}). We skip the parameters $f, b, c, d$ if they are clear from the context.

We treat a multivector corpus as a text document for which the labels are words and the walks are sentences. Next, we apply NLP methods to the document. We emphasize that the methods we use does not depend on the order of sentences, so we can take any order in the $\bigoplus$ notation. On the other hand, the skip-gram model generates word contexts using the sliding window technique. Hence, the order of words in a sequence is important.

\subsection{Encoding}
The {\em distributional hypothesis}~\cite{harris_distributional_1954} in linguistics says that it is possible to state a linguistic structure in terms of patterns of co-occurrences, i.e. words with similar meaning occur in the same context. It is the main idea behind representing words as elements of a vector space. Neural network models~\cite{bengio_neural_2006, bojanowski_enriching_2017, mikolov_efficient_2013, mikolov_distributed_2013} allow us to find an encoding of words from a large text corpus. The intuition behind the vector space elements is that the distance between similar words is small, and the norm of a word encoding is proportional to its importance in the corpus.
In particular, we apply the methods to our artificial multivector corpus introduced in Section~\ref{sec:corpus}.

Recall that we use the {\em Continuous Skip-gram Model}~\cite{mikolov_efficient_2013}
to analyze text documents structure and to find $\Phi$, such that it minimize the following log probability:
\begin{equation}
  \label{eq:prob_max}
  \underset{\Phi}{\text{minimize}} - \log P( \{l_{w_{i-w}}, \ldots, l_{w_{i+w}}\} \setminus \{ l_{w_i} \} \mid \Phi(l_{w_i}) ).
\end{equation}
The model is a shallow neural network trained to predict words within a range before and after the current word. The main parameters for the algorithm are:  {\em window size} $w$ - number of words around current word, {\em encoding dimension} $D$. Intuitively, the parameter $w$ carries examined influence of a word to the meaning of a sentence. The parameter $D$ controls the number of linguistic features learned by the model. From a trained network we extract a $D$ dimensional representation of words, denoted by $\Phi^D$.
For natural languages the parameters values typically are: $w=5$, $D=300$.
In our applications usually much smaller dimension is enough. In the sequel we show, that low dimensional encoding may contain useful information.

In our context $\{l_{w_i}\}_i$ is a sequence of labels in a random walk in a graph $G_\cV$. It is worth to note that, intuitively, labels represent local structures of the dynamics around multivectors and the random walks represent trajectories. For instance we can consider random walks generated by a constant flow and a spiral flow close to a fixed point. We expect that the encodings of multivectors from the two groups should be distinguishable.
It is because the local structures of multivectors on the spiral are different and richer than these on the constant flow. It means that for a randomly chosen label a trained Skip-gram model should be able to predict to which group the label belongs.

The goal of recently developed NLP methods is to find meaningful words encoding. Currently, there is no similar method designed directly for documents. As a workaround a common trick is to use a weighted mean of words encodings as the encoding of a document. In our context, for a multivector field graph $G_{\cV}$, we define $\Phi_w^D(G_{\cV})$ as $ \sum_{v \in V(G_{\cV})} w_v \Phi^D(\cL^{f,b}(v))$, where $f$ and $b$ are fixed and $w_v$ is a weight of $v$, e.g. $\frac{1}{|V(G_{\cV})|}$ or TFIDF of $v$ (term frequency-inverse document frequency \cite{rajaraman_mining_2011}).

\subsection{Implementation details}
\label{sec:impl}
We compute the Skip-gram encoding $\Phi$ using the FastText~\cite{bojanowski_enriching_2017} library. As we mentioned earlier, we cannot use sub-words ($n$-grams) in the training phase, so the parameters \texttt{minn} and  \texttt{maxn} are set to $0$. For the examples presented in this paper, if not stated otherwise,  we set: learning rate to $0.01$, size of the context window to $5$, dimension of word vectors to $2$.
The number of epochs used for training depends on the size of the corpus. We use $1000$ epoches for small corpuses in  Sections~\ref{sec:ex_orbit}~and~\ref{sec:ex_two_orbits}, and $5$ epochs for large corpuses in  Sections~\ref{sec:ex_pp}~and~\ref{sec:ex_ts}.


We also noticed that rare words appear close to the boundary of the sampled region. It is because in that area the neighborhood of a multivector depends more on its location than on the vector field structure. We skip such multivectors by checking their distance to the region boundary. The outcome of this simplification is shown in Figure~\ref{fig:ex_orbit_mvf_colored}, where the area close to the boundary is white, because labels of the multivectors are not in the corpus, so $\Phi$ maps them to zero.

\section{Examples}
\label{sec:ex}
In this section we present the methods in action.
First we show toy examples which illustrate a qualitative properties of the encodings. Later we present applications to a series of dynamical systems analysis. Our examples are simple enough to visualize the encodings.
However, in real world applications higher values of the encoding dimension ($D$) and the labeling level ($\cL^{b,f}$)may be required. Then, one may need to create a pipeline, where the output of our method is only an intermediate step. We want to keep the paper simple and more complicated applications, e.g. turbulences analysis or solar flare classifications, we are going to present in a sequel paper. The presented examples illustrate that the methods are able to automatically extract meaningful features from dynamical systems.

We begin with a setup for computations. For a product of $k$ intervals $ I = [I^-_1, I^+_1] \times \ldots \times [I^-_k, I^+_k] \subset \RR^k$ and a set of natural numbers $N = \{N_1, \ldots, N_k \} \subset \NN_+$, we define an $(I, N)$ {\em regular grid of points}, denoted by $\Rgrid(I, N)$, as a set of points $\{ (x_1, \ldots, x_k) \in I \mid x_i = I^-_i + \frac{ (I^+_i - I^-_i) j }{N_i} \text{ for } j \in [0, N_i] \subset \NN  \}$.

In the context of this paper we assume a family $\gV$ of combinatorial multivector fields is given. In order to present the examples we recall a possible way to construct a  combinatorial multivector field from a cloud of vectors.
Let $K$ be a simplicial complex with vertices in a cloud of points $\{p_i\mid i=1,2,\ldots, n\}\subset \mathbb{R}^d$
and the associated cloud of vectors $\{   \vec{v_i} \mid i=1,2,\ldots, n\}\subset\RR^d$ such that vector $\vec{v_i}$ originates from point $p_i$.
To construct a  \cmf{} on $K$ one can use the algorithm $\mathrm{CVCMF}$~\cite[Table 1]{dey_persistent_2019} (called here $\mathrm{CVCMF} v1$).
The algorithm requires an angular parameter $\alpha$.
We do not analyze the impact of the parameter here. We also have a parameter-less version of the algorithm (called here $\mathrm{CVCMF} v2$). We show results obtained with the old and the new algorithms, however the new one is not published yet.

In the below examples we plot $D$ dimensional encodings of the words obtained with the FastText~\cite{bojanowski_enriching_2017} implementation. On the plots we observe cone-shaped points distribution. It suggest that it should be enough to use the encoding dimension parameter value equal to $D - 1$. However, this is an outcome of the negative sampling training algorithm and this is a phenomenon described in~\cite{mimno_strange_2017}.

\subsection{Example: Orbit}
\label{sec:ex_orbit}
\begin{figure*}[!htbp]
  \centering
  \subfloat[Vector field with the periodic orbit and the stationary point marked.]{
    \includegraphics[width=.3\textwidth]{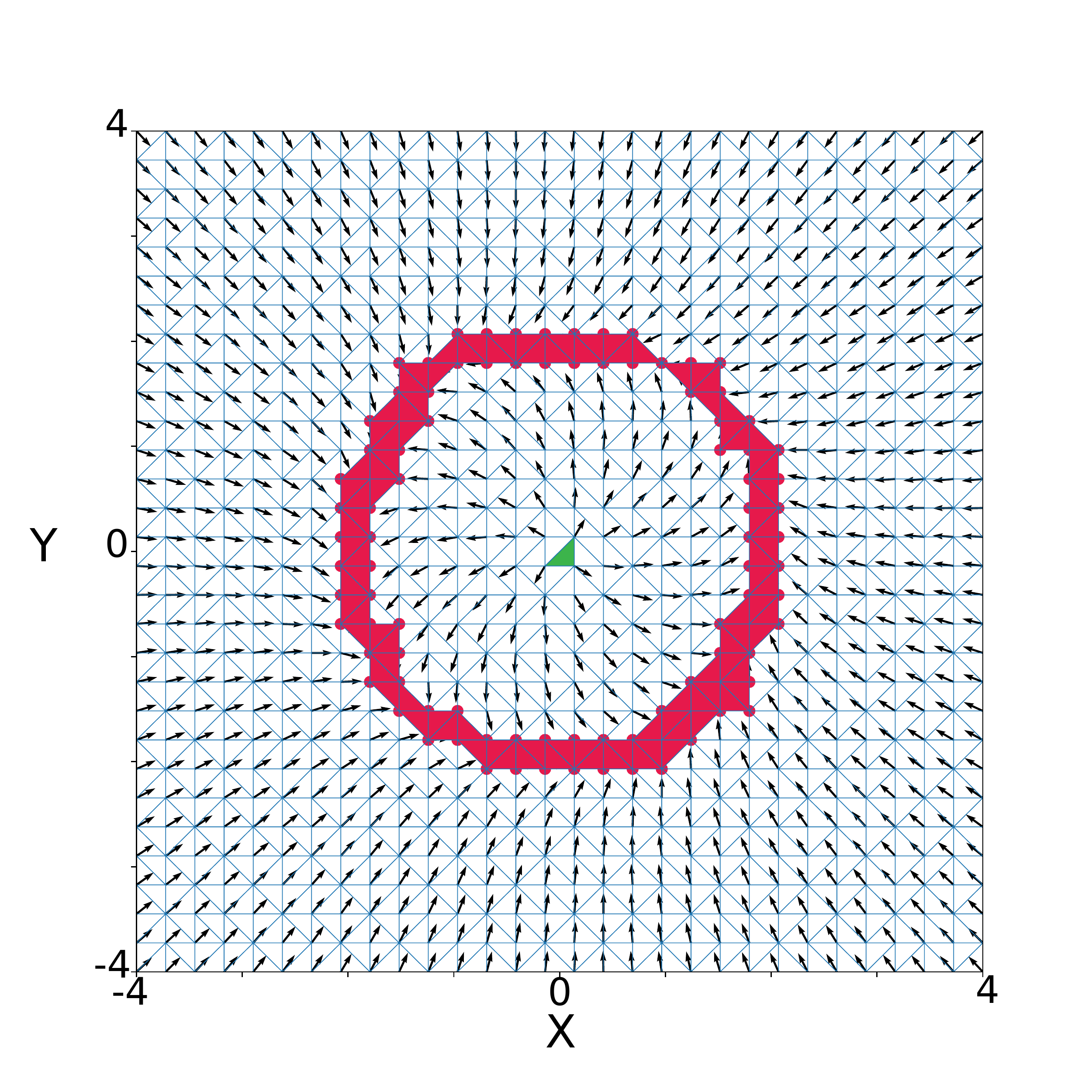}
    \label{fig:ex_orbit_morse}
  }
  \hfill
  \subfloat[2D encoding of the multivectors.]{
    \includegraphics[width=.3\textwidth]{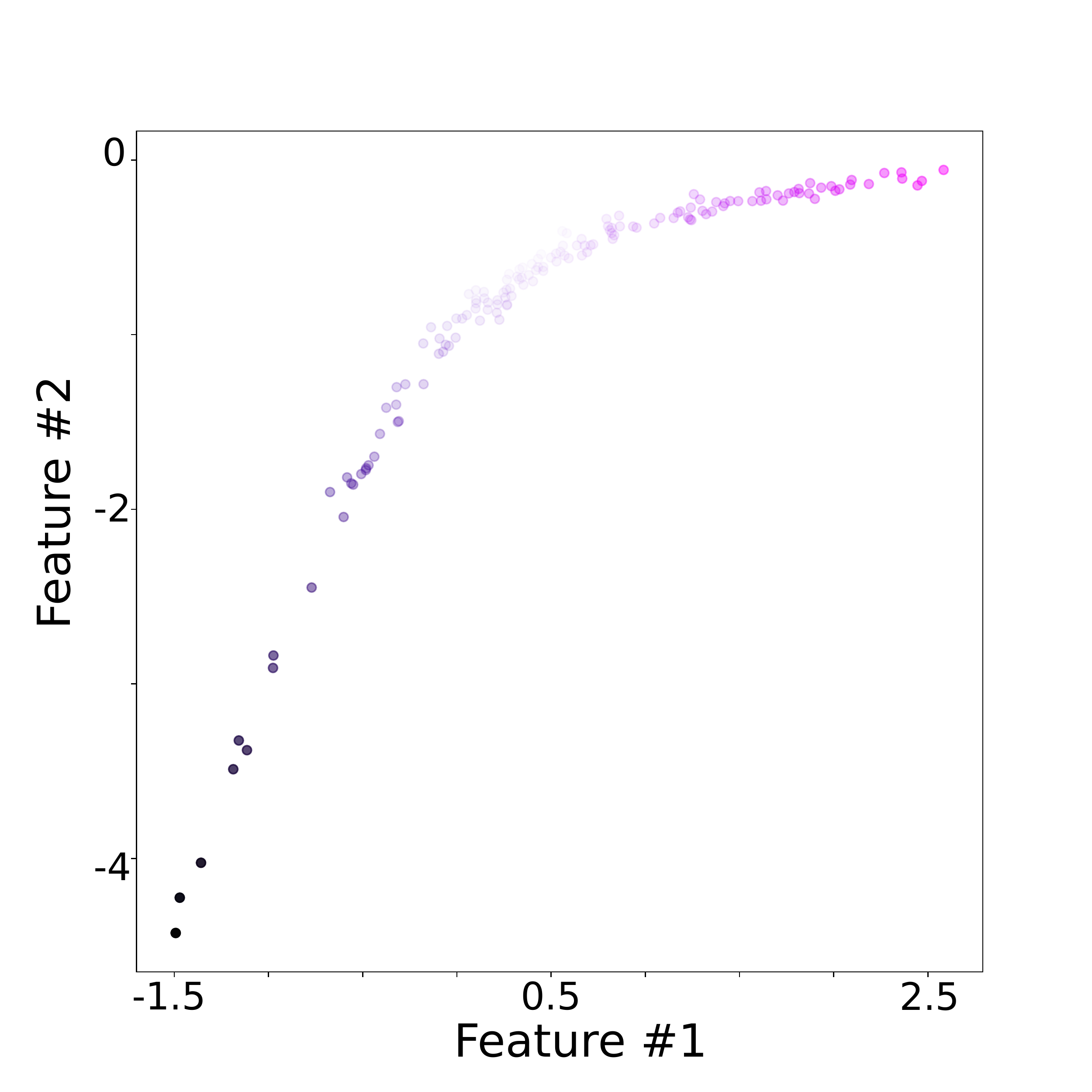}
    \label{fig:ex_orbit_emb}
  }
  \hfill
  \subfloat[Multivectors colored using colors of the encodings.]{
    \includegraphics[width=.3\textwidth]{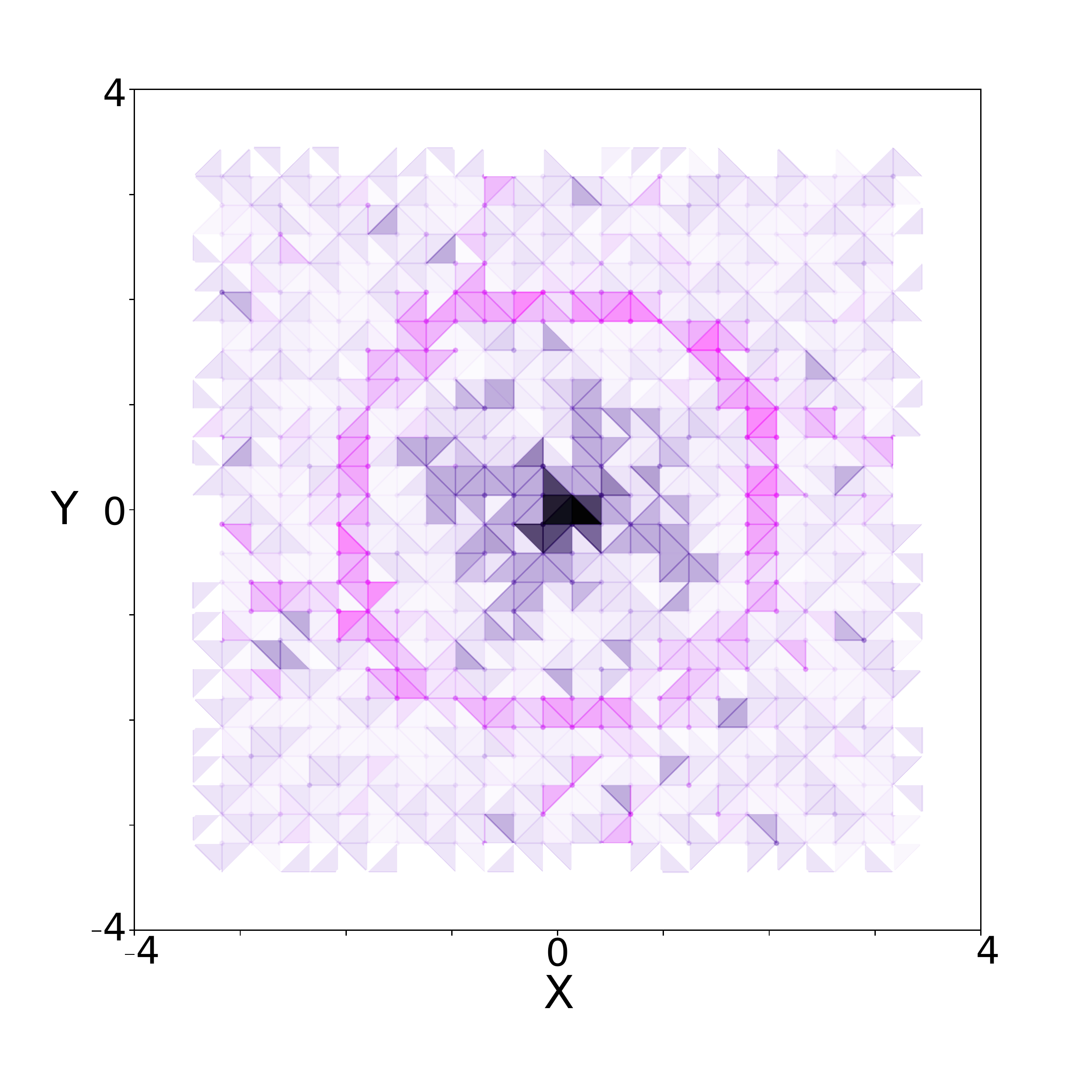}
    \label{fig:ex_orbit_mvf_colored}
  }
  
  \caption{An example of a multivector field encoding for a single orbit and a repelling point.}
  \label{fig:ex_orbit}
\end{figure*}

The main goal of this example is to show features extracted by our method on a simple dynamical system.
Consider a system given by the following equation:
\begin{equation}
  \diff{x}{t} = -y + x (4 - x^2 - y^2) ,
  \qquad
  \diff{y}{t} = -x + y (4 - x^2 - y^2).
\end{equation}
In the system phase space we  observe a repelling stationary point at $(0,0)$ and an attracting periodic orbit with center at $(0,0)$ and radius $2$.
We can observe this
in a combinatorial dynamical system constructed from a finite sample of the vector field
using methods described in~\cite{dey_persistent_2019, mrozek_conleymorseforman_2017}. To achieve that we build a Delaunay triangulation $K$ of a regular grid $\Rgrid([-4,4] \times [-4, 4], \{30, 30\})$.
The triangulation contains $1682$ triangles and $5163$ simplices.
To construct a \cmf{} $\cV$ of the triangulation and vectors originates from its vertices we use the algorithm $\mathrm{CVCMF} v1$~\cite[Table 1]{dey_persistent_2019} with the parameter $\alpha=0$. The triangulation $K$
with the discretized periodic orbit and the repelling point of the system are presented in Figure~\ref{fig:ex_orbit_morse}.

In the next step we label vertices of the graph $G_{\cV}$ using labels at level $(1,1)$, getting $153$ distinct labels. Using the labeled graph we generate $(1, 1, 2, 10)$-corpus for which we obtain the Skip-gram encoding $\Phi$ with parameters $w=5$ and $D=2$. The encodings are plotted in Figure~\ref{fig:ex_orbit_emb}, where each dot represents an unique word in the corpus (label in the graph). We use RGBA color space, where the alpha channel of a point is proportional to its norm. 

In Figure~\ref{fig:ex_orbit_mvf_colored}, we show colored multivectors, where each multivector $v$ gets color of $\Phi(\cL(v))$. We notice that the encoding $\Phi$ distinguish three regions of the phase space with following behaviors: close to the the orbit, around the repelling point, outside the orbit.

\subsection{Example: Nested orbits}
\label{sec:ex_two_orbits}
Consider a system given by the following equation:
\begin{align}
  \label{eq:two_orbits}
  \begin{split}
  \diff{x}{t} & =  -0.3y((x^2 + y^2 - 1) - (x^2 + y^2 - 1)^2) - x(3 - 6(x^2 + y^2 - 1) + (x^2+y^2 - 1)^2),\\
  \diff{y}{t} & =  0.3x((x^2 + y^2 -1) - (x^2 + y^2 - 1)^2) - y(3 - 6(x^2 + y^2 - 1) + (x^2+y^2 - 1)^2).
  \end{split}
\end{align}
In the system phase space we observe an attracting stationary point at $(0,0)$ and two periodic orbits with centers at $(0,0)$.

In this section we show features extracted by our method from a sampled dynamical system given by~\eqref{eq:two_orbits}.
As in Section~\ref{sec:ex_orbit} we build a combinatorial dynamical system using a Delaunay triangulation $K$ of a regular grid $\Rgrid([-4,4] \times [-4, 4], \{50, 50\})$.
To construct a \cmf{} $\cV$ of the triangulation and vectors originates from its vertices we use the algorithm $\mathrm{CVCMF} v2$. Here we use the labels at level $(2,2)$ and the encoding dimension $D=3$. With lower values of the parameters we cannot distinguish multivectors on the orbits and the stationary point.

\begin{figure*}[!htbp]
  \centering
  \subfloat{
    \includegraphics[width=.3\textwidth]{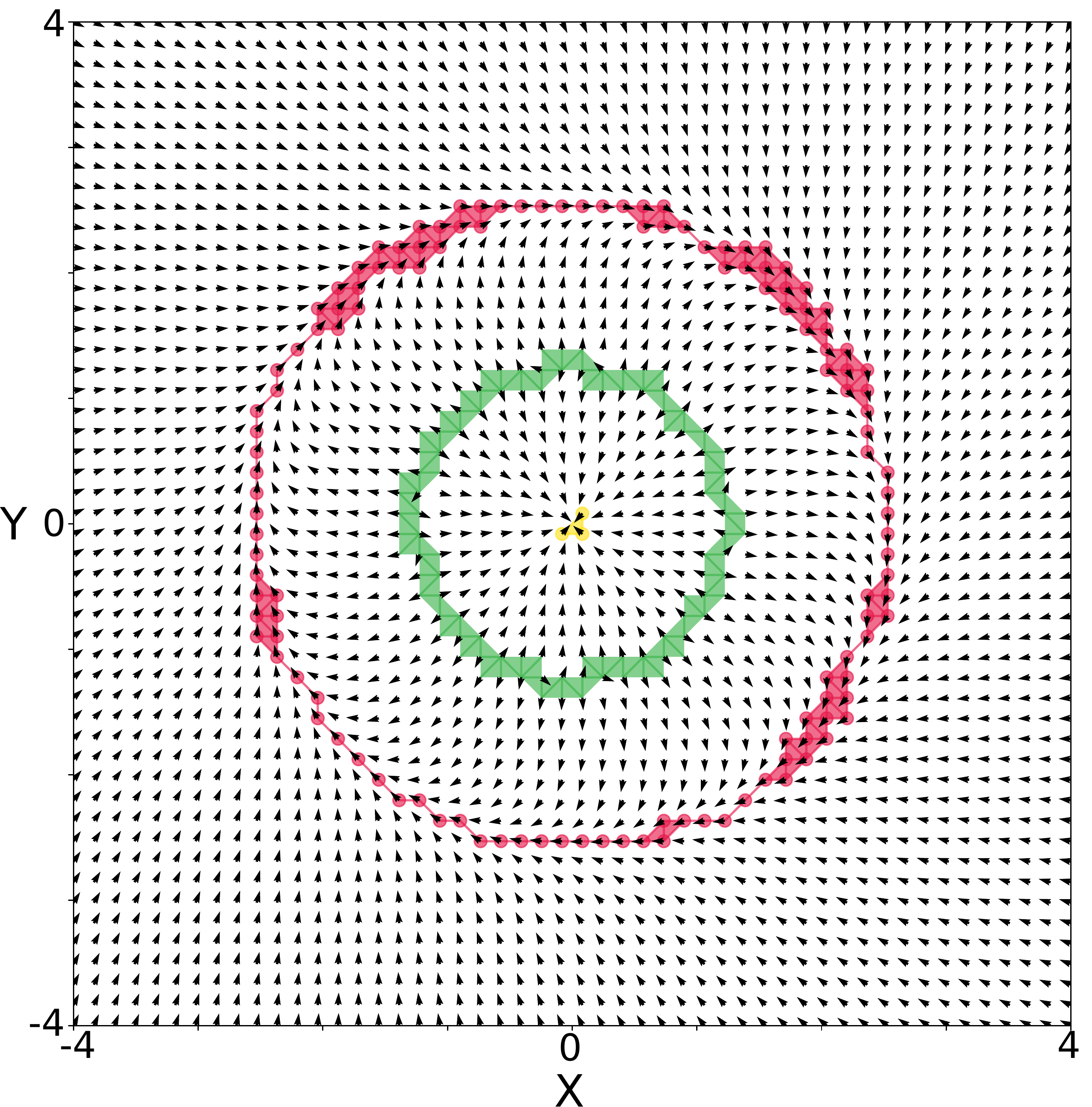}
    \label{fig:ex_two_orbits_morse}
  }
  \hfill
  \subfloat{
    \includegraphics[width=.3\textwidth]{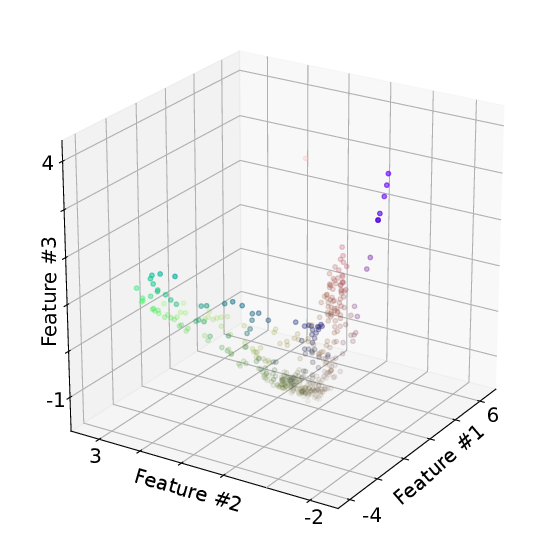}
    \label{fig:ex_two_orbits_emb}
  }    
  \hfill
  \subfloat{
    \includegraphics[width=.3\textwidth]{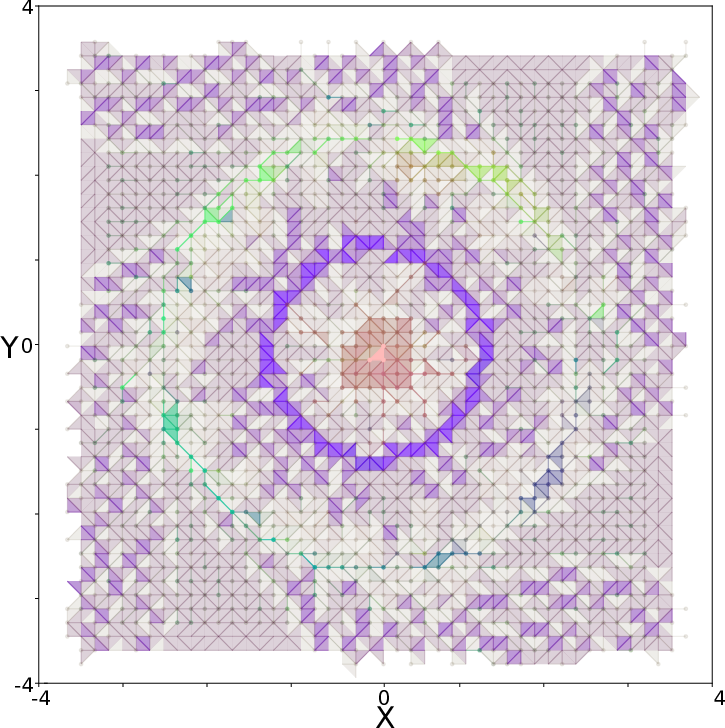}
    \label{fig:ex_two_orbits_emb}
  }  
  \caption{A visualization for the system generated by~\eqref{eq:two_orbits} using $(2,2,1,30)$-corpus and $w=10$:
    (left) the vector field with two periodic orbits and an attracting point marked; (middle) $3$D encodings of the multivectors (RGB colors given by a point coordinates and the alpha channel is proportional to the point norm); (right)  the multivectors field colored using the encodings.
  }
  \label{fig:ex_two_orbits}
\end{figure*}

\begin{figure*}[!htbp]
  \centering
  \subfloat[$d = 10$, $w=2$.]{
    \includegraphics[width=.3\textwidth]{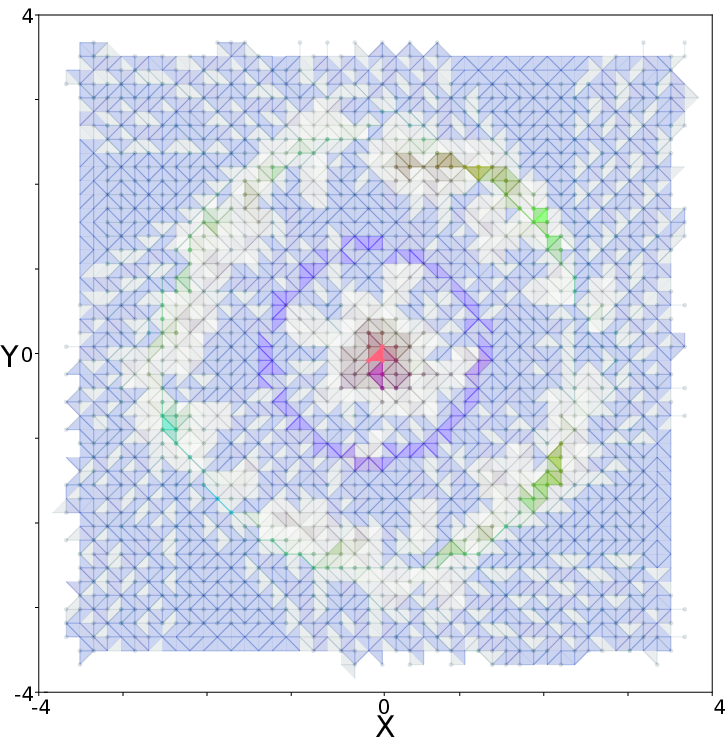}
  }
  \hfill
  \subfloat[$d = 10$, $w=5$.]{
    \includegraphics[width=.3\textwidth]{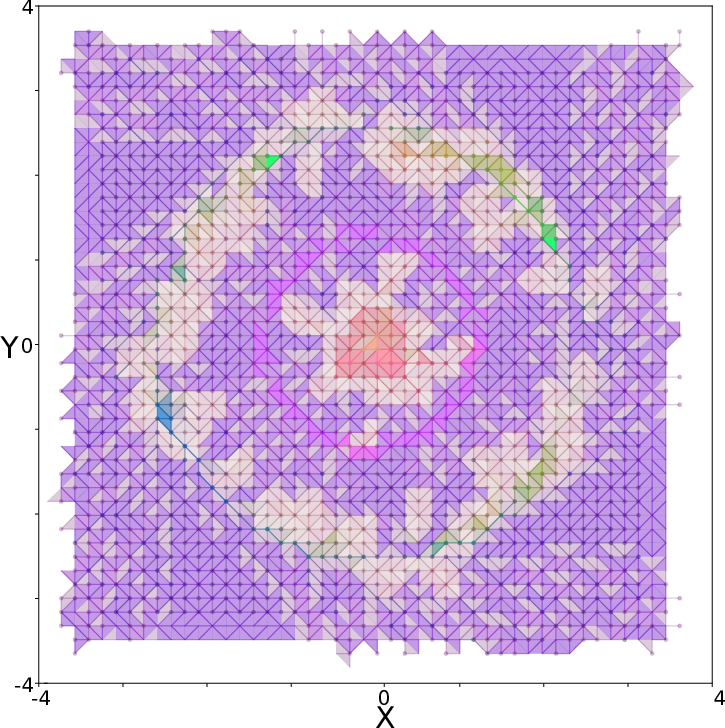}
  }
  \hfill
  \subfloat[$d = 10$, $w=10$.]{
    \includegraphics[width=.3\textwidth]{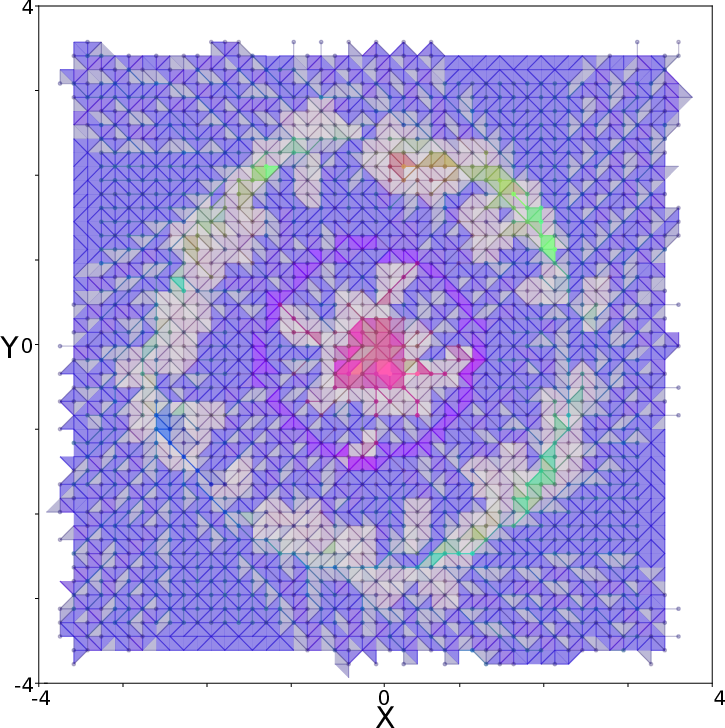}
  }

  \subfloat[$d = 20$, $w=2$.]{
    \includegraphics[width=.3\textwidth]{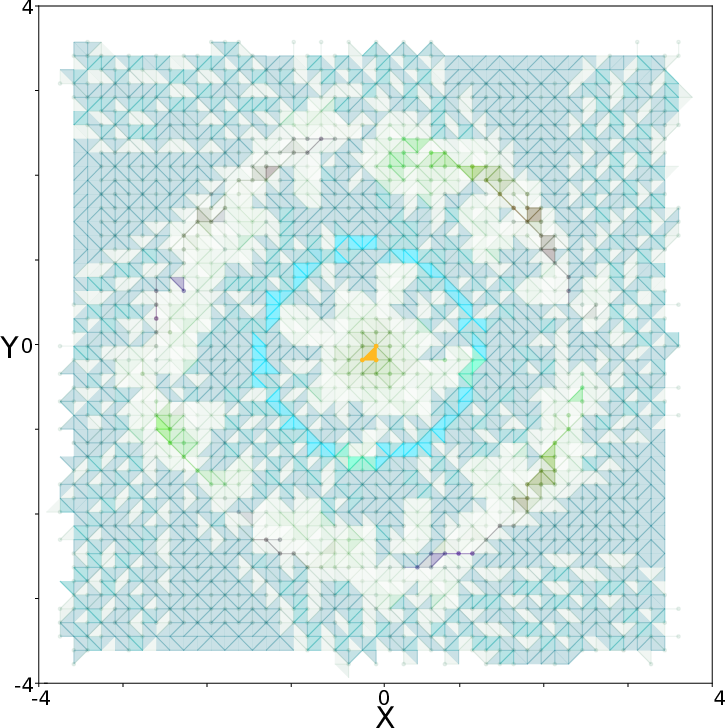}
  }
  \hfill
  \subfloat[$d = 20$, $w=5$.]{
    \includegraphics[width=.3\textwidth]{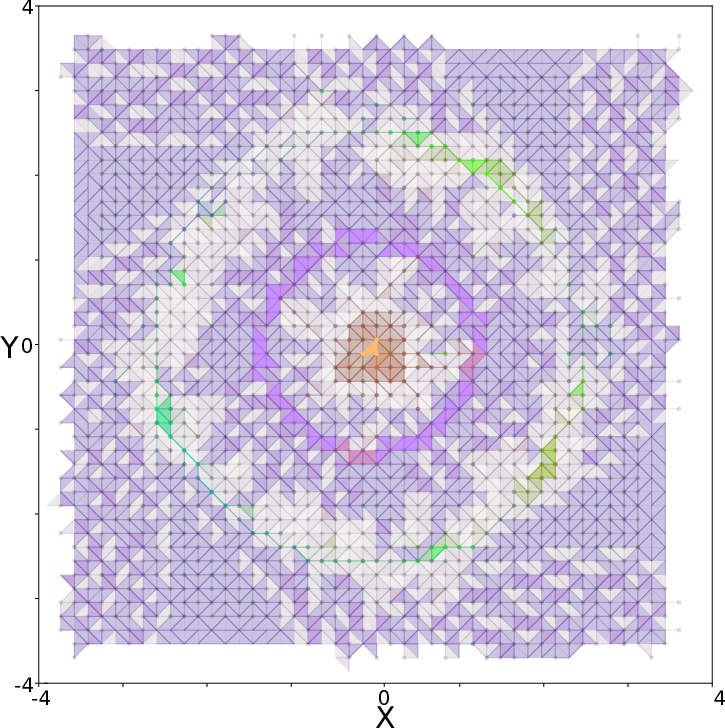}
  }
  \hfill
  \subfloat[$d = 20$, $w=10$.]{
    \includegraphics[width=.3\textwidth]{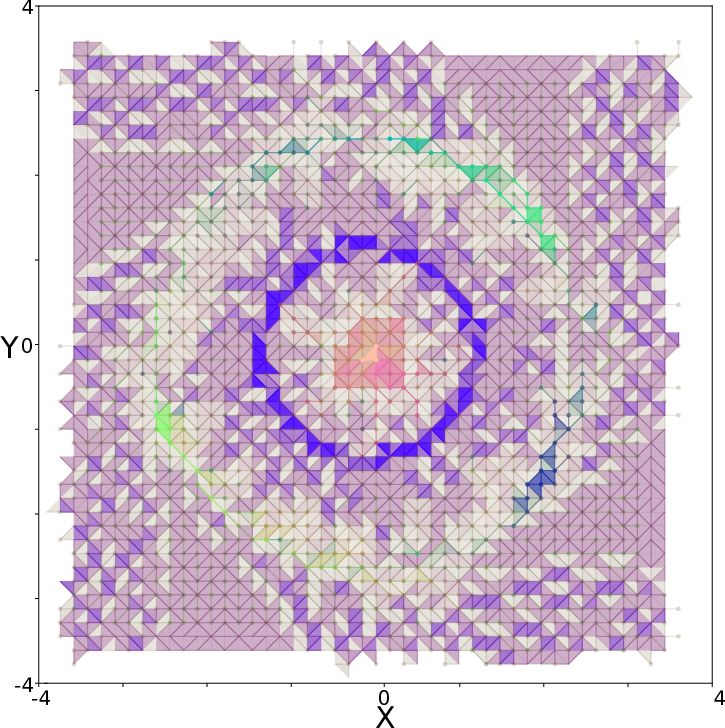}
  }

  \subfloat[$d = 30$, $w=2$.]{
    \includegraphics[width=.3\textwidth]{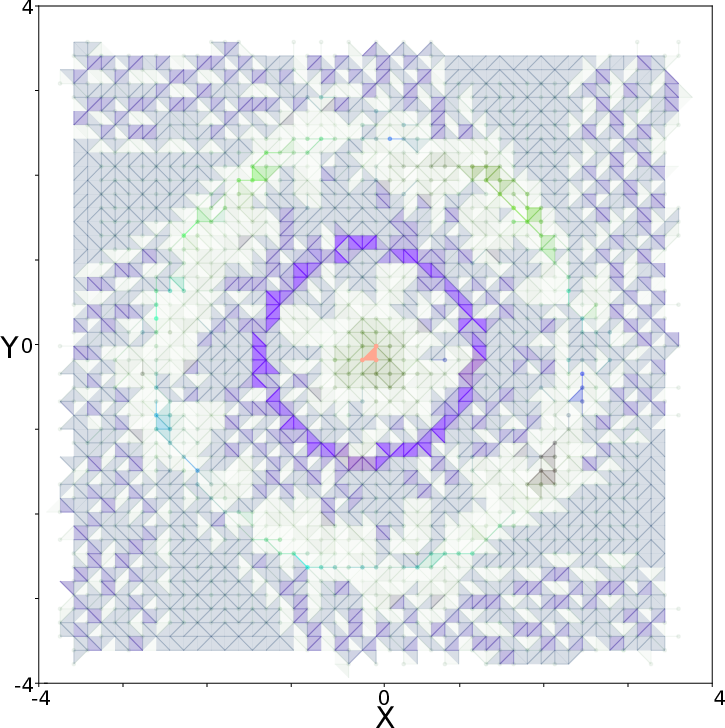}
  }
  \hfill
  \subfloat[$d = 30$, $w=5$.]{
    \includegraphics[width=.3\textwidth]{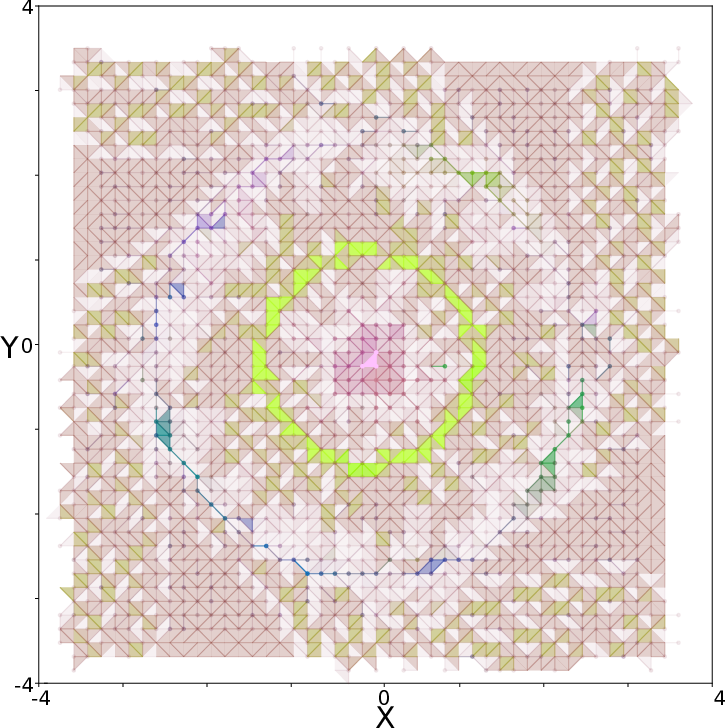}
  }
  \hfill
  \subfloat[$d = 30$, $w=10$.]{
    \includegraphics[width=.3\textwidth]{two_orbits-dim_3-l_2_2-d_30_30-w_10.png}
  }
  
  \caption{
    Visual evaluation of the method for the system generated by $\eqref{eq:two_orbits}$. We use following settings: $D=3$,  $w \in \{2, 5, 10\}$, $(2, 2, 1, d)$-corpuses for $d \in \{10, 20, 30\}$.  For each test case we compute its own Skip-gram model and the $3$D encodings. We use a multivector encoding as its RGBA color (see Figure~\ref{fig:ex_two_orbits}).  We observe that the triangle with the attracting point at $(0,0)$ is not clearly visible for $d=10$, and the outer orbit merges with its neighbors for $w=2$.}
  \label{fig:ex_two_orbits_by_corpuses}
\end{figure*}

In Figure~\ref{fig:ex_two_orbits} we show the triangulation $K$, and the \cmf{} $\cV$ colored according to the encoding of the multivectors. Afterwards, in Figure~\ref{fig:ex_two_orbits_by_corpuses} we show the influence of $w$ and $d$ parameters. In conclusion, for values $d \ge 20$ and $w \ge 5$ we can observe a clear difference between the encodings for orbits and the stationary point.

\subsection{Example: Prey-predator}
\label{sec:ex_pp}
In this section we use the presented methods to analyze a series of dynamical systems. We also show the influence of  the labeling levels and the $\mathrm{CVCMF}$ algorithm variants.
We consider a prey-predator model~\cite{ghosh_prey-predator_2017} given in the following form:
\begin{equation}
  \diff{x}{t} =  x(1 - \frac{x}{\gamma}) - \frac{(1 - c)xy}{1 + \alpha\zeta + x},
  \qquad
  \diff{y}{t} = \frac{\beta[(1-c)x + \zeta]y}{1 + \alpha\zeta + x} - \delta y
  \label{eq:pp-model}
\end{equation}
where $x$ and $y$ denote the biomass of prey and predator respectively, and $\alpha,\beta,c,\delta,\gamma,\zeta$ are parameters.
We investigate sampled vector spaces obtained from simulations of the model given by~\eqref{eq:pp-model}.
In particular, we are interested in qualitative behavior of the model, where
$\alpha \in [0, 2]$, $c \in [0, 0.45]$,
$\beta=0.15, \delta=0.08, \gamma = 4, \zeta=0.2$. For varying $\alpha$ and $c$ we denote the model by $P(\alpha, c)$.

\begin{figure}[!htbp]
  \centering
  \includegraphics[scale=0.3]{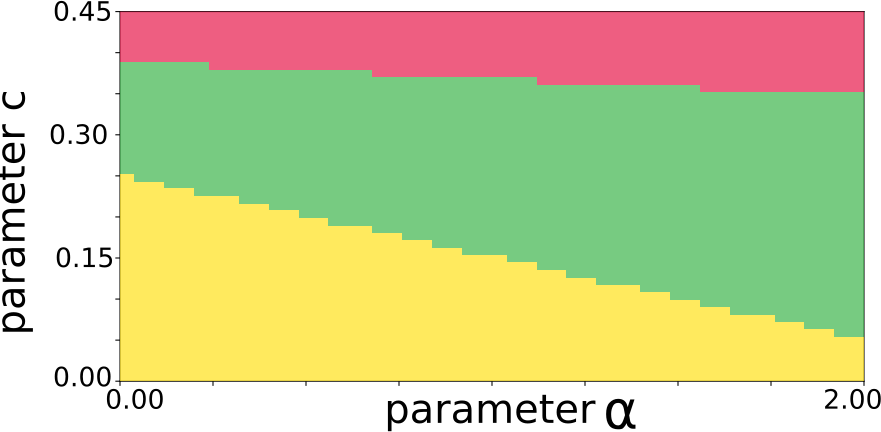}
  \caption{Qualitative behavior of the model~\eqref{eq:pp-model} according to~\cite{ghosh_prey-predator_2017}. The parameter plane $(\alpha, c) \subset [0, 2]\times[0, 0.45]$ is divided into three regions: oscillatory coexistence (yellow), stable coexistence (green), predator extinction (red).}
  \label{fig:pp-params}
\end{figure}

In Figure~\ref{fig:pp-params} we show a decomposition of the parameter plane $(\alpha, c)$ obtained analytically in~\cite{ghosh_prey-predator_2017}.
Our goal is to show a correlation between the model dynamics and multivectors encoding $\Phi$.

Let $K$ be a Delaunay triangulation of a regular grid $\Rgrid([0, 6]\times[0.4, 2.4], \{100, 100\})$. Let $\cV_{\alpha, c}^{\theta} $ be a \cmf{} of $K$ computed using
the $\mathrm{CVCMF}$ algorithm (see beginning of  Section~\ref{sec:ex}) called with $K$ and vector field sampled from the prey-predator model $P(\alpha, c)$. The value of $\theta$ controls the $\mathrm{CVCMF}$ algorithm version, namely:
\begin{itemize}
\item[-] if $\theta \in [0, 2\pi]$, then we use $\mathrm{CVCMF} v1$ with its angular parameter equals to $\theta$,
\item[-] if $\theta = \emptyset$, then we use $\mathrm{CVCMF} v2$.
\end{itemize}

\begin{figure*}[!htbp]
  \centering
  \subfloat{
    \includegraphics[width=.28\textwidth]{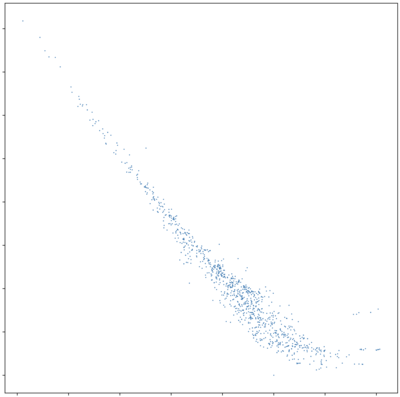}
    \label{fig:ex_pp_L1_gready_labels}
  }
  \hfill
  \subfloat{
    \includegraphics[width=.28\textwidth]{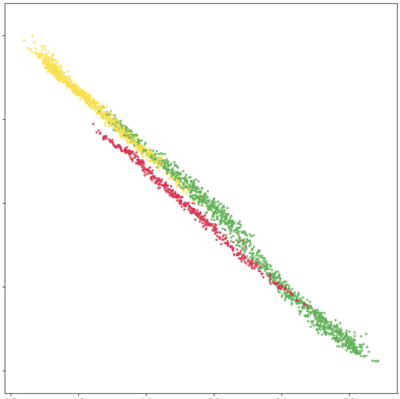}
    \label{fig:ex_pp_L1_gready_mean}
  }
  \hfill
  \subfloat{
    \includegraphics[width=.28\textwidth]{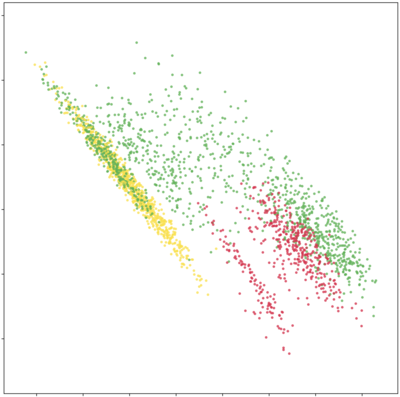}
    \label{fig:ex_pp_L1_gready_tfidf}
  }
  \vspace{-0.4cm}
  \newline
  \subfloat{
    \includegraphics[width=.28\textwidth]{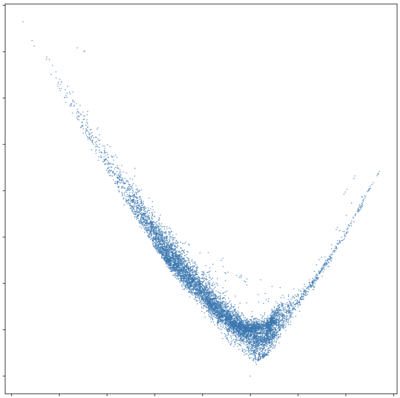}
    \label{fig:ex_pp_L1_gready_labels}
  }
  \hfill
  \subfloat{
    \includegraphics[width=.28\textwidth]{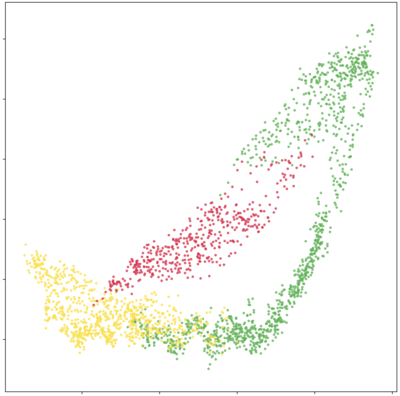}
    \label{fig:ex_pp_L1_gready_mean}
  }
  \hfill
  \subfloat{
    \includegraphics[width=.28\textwidth]{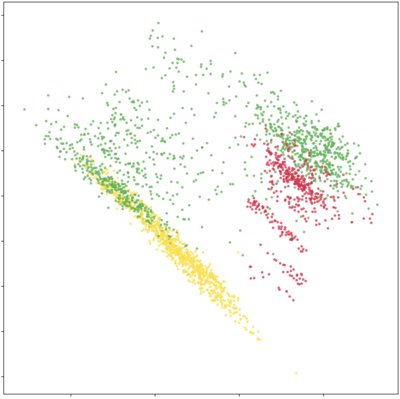}
    \label{fig:ex_pp_L1_gready_tfidf}
  }
  \vspace{-0.4cm}
  \newline
  \subfloat{
    \includegraphics[width=.28\textwidth]{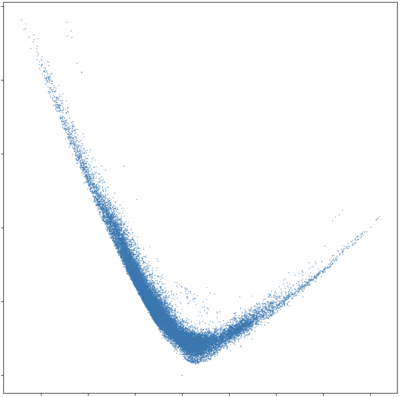}
    \label{fig:ex_pp_L1_gready_labels}
  }
  \hfill
  \subfloat{
    \includegraphics[width=.28\textwidth]{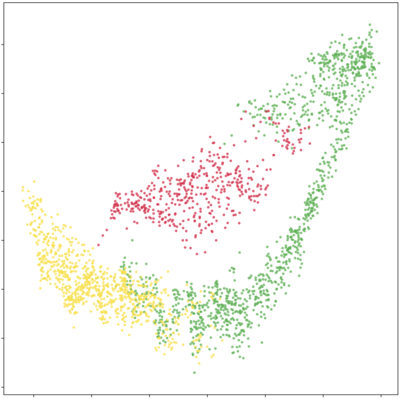}
    \label{fig:ex_pp_L1_gready_mean}
  }
  \hfill
  \subfloat{
    \includegraphics[width=.28\textwidth]{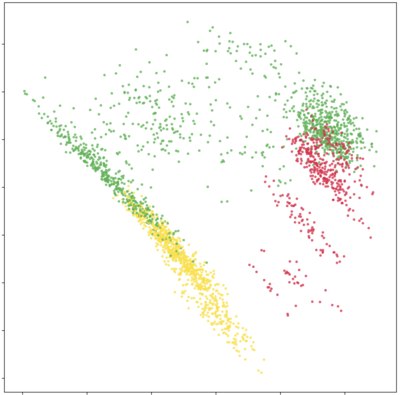}
    \label{fig:ex_pp_L1_gready_tfidf}
  }
  \vspace{-0.4cm}
  \newline
  \subfloat{
    \includegraphics[width=.28\textwidth]{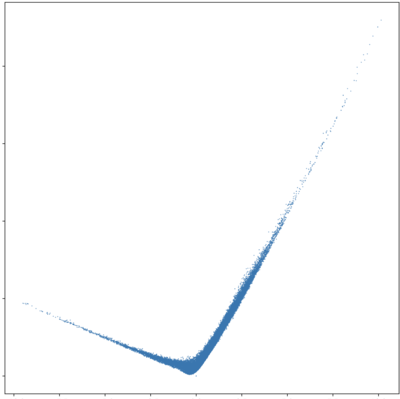}
    \label{fig:ex_pp_L1_gready_labels}
  }
  \hfill
  \subfloat{
    \includegraphics[width=.28\textwidth]{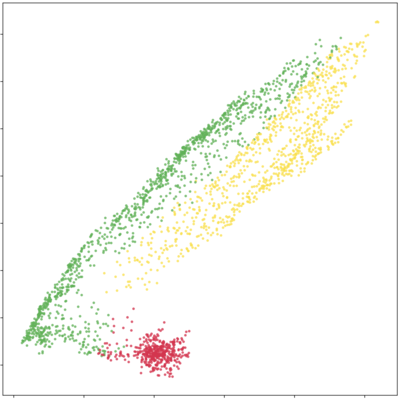}
    \label{fig:ex_pp_L1_gready_mean}
  }
  \hfill
  \subfloat{
    \includegraphics[width=.28\textwidth]{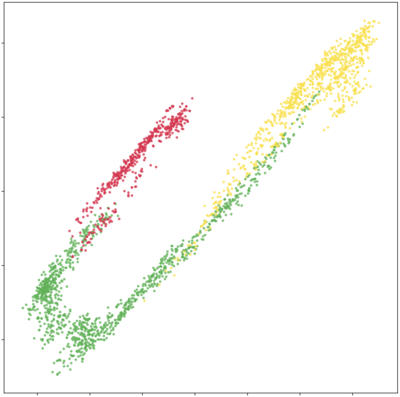}
    \label{fig:ex_pp_L1_gready_tfidf}
  }
  \caption{$2$-dimensional encodings for the prey-predator multivector corpuses. Each row represents a corpus: $\cC_{I}, \cC_{II}, \cC_{III}, \cC_{IV}$ (top-down). First column represents encodings of labels: for each distinct label $l$ in a multivector corpus there is a dot at position $\Phi^2(l)$. Second and third columns represents encodings of graphs: for each $\cV = \cV^\theta_{\alpha,c} \in \gV^\theta$ there is a dot at position $\Phi_w^2(G_\cV)$, where $w$ is the mean (second column) or TFIDF (third column) weighting. Colors of the dots correspond to the colors of $(\alpha, c)$ in Figure~\ref{fig:pp-params}. We are interested in the points distributions, so we skip values on the plots axes.}
  \label{fig:ex_pp}
\end{figure*}

Let $\gV^\theta$ be a family of combinatorial multivector fields
\[
\{ \cV^\theta_{\alpha, c} \mid (\alpha, c) \in \Rgrid([0,2]\times[0, 0.45], \{50, 50\}) \},\text{ for some fixed } \theta.
\]
We train the Skip-gram neural network on four corpuses described in Table~\ref{tab:ex_pp_tests} using the FastText~\cite{bojanowski_enriching_2017} library (see Section~\ref{sec:impl}) and obtain $2$-dimensional encodings for each of them. For each corpus we present in Figure~\ref{fig:ex_pp} the encoding of each word and the encoding of each multivector field graph. The pictures suggest that it should be possible to distinguish more types of the prey-predator model dynamics, i.e. the structure of the green subspace does not arrange into a blob of points. We also observe that there is no big difference between corpuses $\cC_{II}$ and $\cC_{III}$. The best separation between green and yellow points is visible for $\cC_{IV}$, which suggest that our new non-parameterized algorithm $\mathrm{CVCMF} v2$ finds a better partition of a given complex into multivectors. Further research using machine learning classifiers and higher dimensional encodings are in progress.

\begin{table}
\centering
  \begin{tabular}{|| c V{3} c | c | c||}
 \hline
 coprus variant & \# of distinct words & \# of words in the corpus \\ [0.5ex]
 \hline\hline
 $\cC_{I} = \corpus{1}{1}{5}{10}{\gV^{36^\circ}}$ & $965$ & $3877  \cdot 10^6$ \\
 \hline
 $\cC_{II} = \corpus{2}{2}{5}{10}{\gV^{36^\circ}}$ & $7979$ & $3665  \cdot 10^6$ \\
 \hline
 $\cC_{III} = \corpus{3}{3}{5}{10}{\gV^{36^\circ}}$ & $28124$ & $3635  \cdot 10^6$ \\
 \hline
 $\cC_{IV} = \corpus{3}{3}{5}{10}{\gV^{\emptyset}}$ & $96589$ & $4825 \cdot 10^6$ \\ [1ex]
 \hline
  \end{tabular}
  \caption{Multivector corpuses tested for the prey-predator model. Columns: the number of distinct words in a corpus and the total number of words in a corpus.}
    \label{tab:ex_pp_tests}
\end{table}

\subsection{Example: Time series data}
\label{sec:ex_ts}
In this section we present a not obvious application of the methods. Namely, we
automatically extract features from time series data sets using an approach inspired by~\cite{wang_imaging_2015}.
The key idea is to transform a time series data into a sampled vector field.

\begin{figure}[!htbp]
  \begin{center}
    \includegraphics[scale=.3]{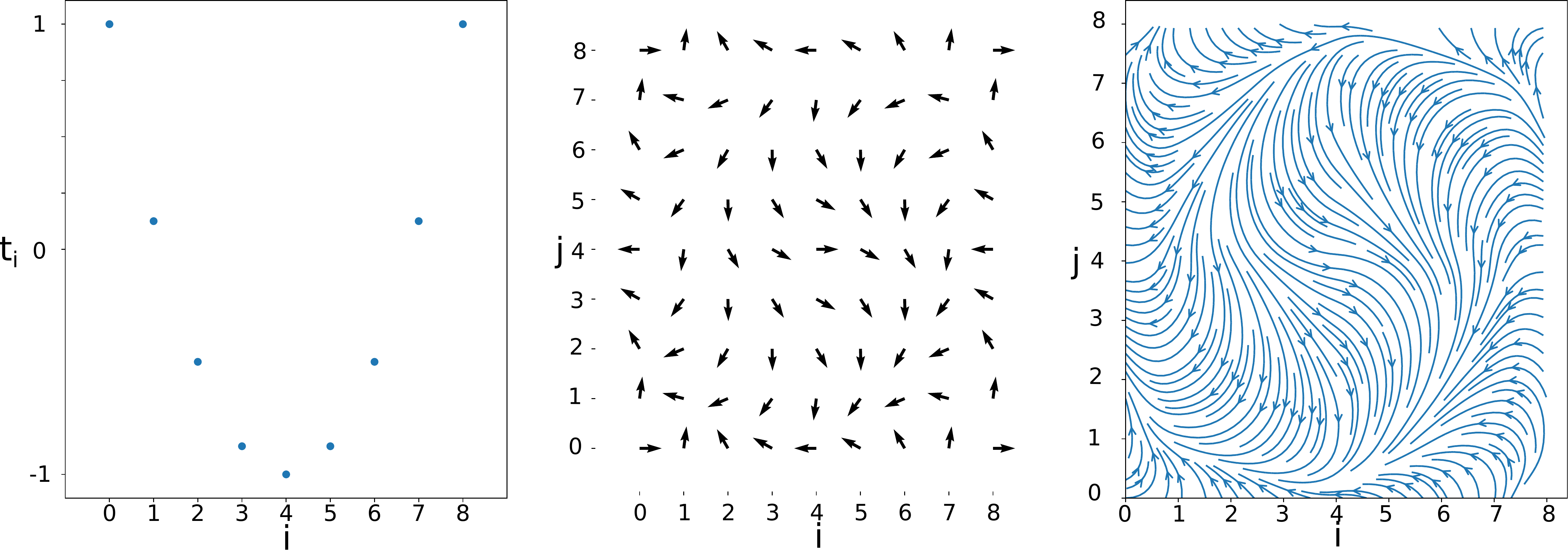}
  \end{center}
  \caption{Left: min-max scaled time series $\cT = \{ (i-4)^2 \}_{i = 0}^{8}$. Middle: on the axes are two copies of the temporal coordinate and on the plot are the $(i,j)$-vectors of $\cT$, for $i,j \in \{ 1,\ldots,8 \}$. Right: a stream plot visualization of the $(i,j)$-vectors.
 It is worth to note that the time series data is rather sparse but yet the vector field looks smooth and it is possible to create an expressive stream plot.}
  \label{fig:vf_1}
\end{figure}

We assume that values of a time series lie in the range $[-1,1]$. Otherwise we can re-scale the values, e.g. using min-max scaling.
Let $\mathcal{T} = \{ t_i \}_{i=0}^{n-1} \subset [-1, 1]$ be a time series. In~\cite{wang_imaging_2015} the authors
introduced the {\em Gramian Summation Angular Field} which is a transformation of $\mathcal{T}$ into a gray-scale $2$D image.
A $(i,j)$-th pixel color is defined as  $\cos(\arccos(t_i) + \arccos(t_j))$.
We propose to use similar technique, but to create vectors. Let $K(\mathcal{T})$ be a Delaunay triangulation of a regular grid $\Rgrid([0, n]\times[0, n], \{n, n\})$, where $n$ is the length of $\mathcal{T}$. Let $\alpha_{i,j} := \arccos(t_i) + \arccos(t_j)$ for each vertex $(i,j)$ of the grid vertex.
At each vertex $(i,j)$ we attach a vector $v_{i,j} := (\cos \alpha_{i,j}, \sin \alpha_{i,j})$.
Of course this is an arbitrary choice, but a further investigation of different possibilities is beyond the scope of this paper.
In Figure~\ref{fig:vf_1} we show a toy example.

As a first validation of the method we use {\em Symbols} dataset from the UEA \& UCR Time Series Classification Repository~\cite{bagnall16bakeoff}. The dataset contains a collection of
time series representing the motion on the x-axis of a hand drawing of a specified shape.
We have $6$ classes of shapes and each time series has length $398$.

\begin{figure}[!htbp]
  \begin{center}
    \includegraphics[scale=0.35]{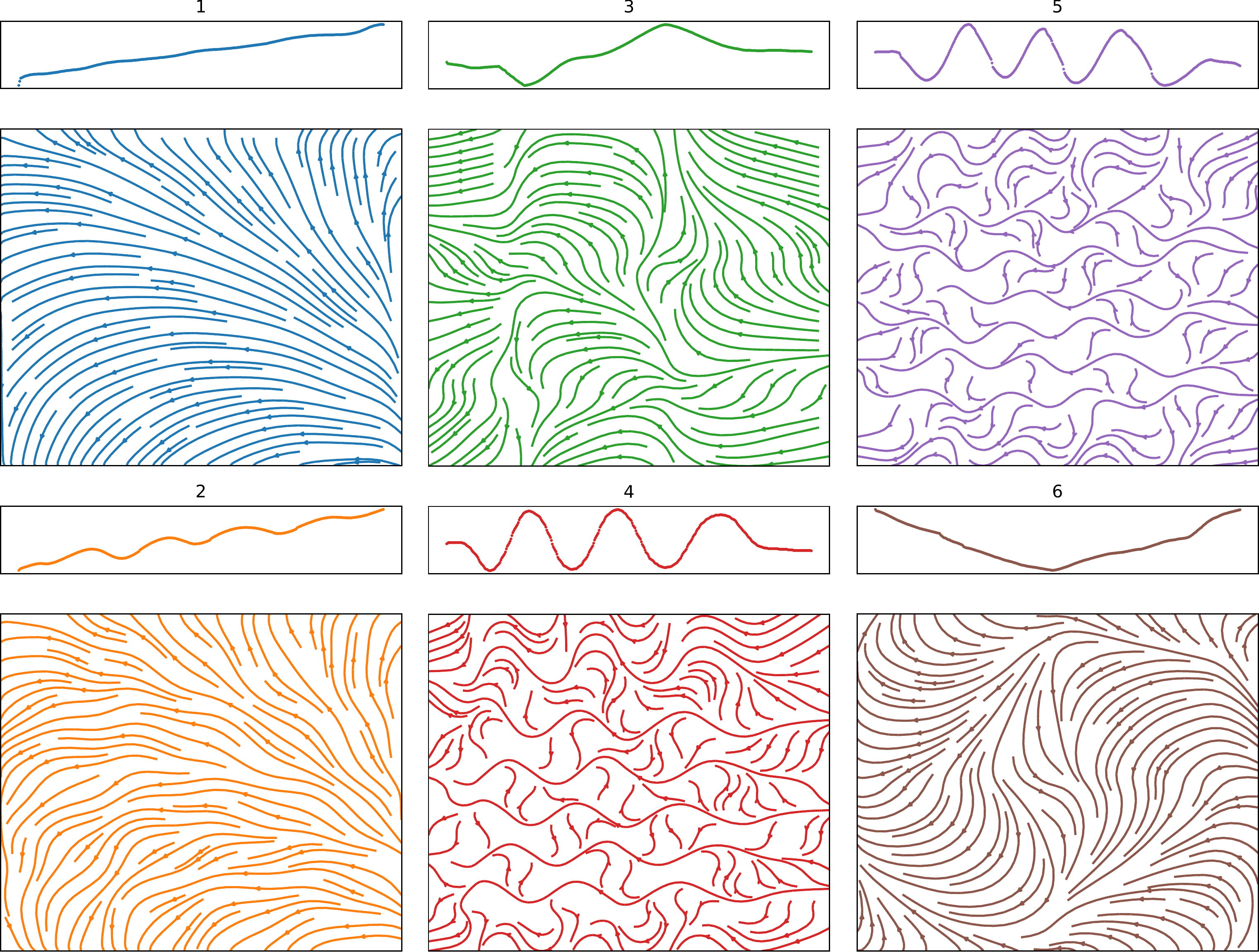}
  \end{center}
  \caption{Time series from the {\em Symbols} data set: one example of a series per class and the stream plots of their vector fields.}
  \label{fig:symbols_ts}
\end{figure}

The dataset is splitted into training set of size $25$ and test set of size $995$. In Figure~\ref{fig:symbols_ts} we show a few examples from the dataset. Note, that the presented vector fields do not have critical points and only the curvature of the flows distinguish them.

\begin{figure}[!htbp]
  \begin{center}
    \includegraphics[scale=0.35]{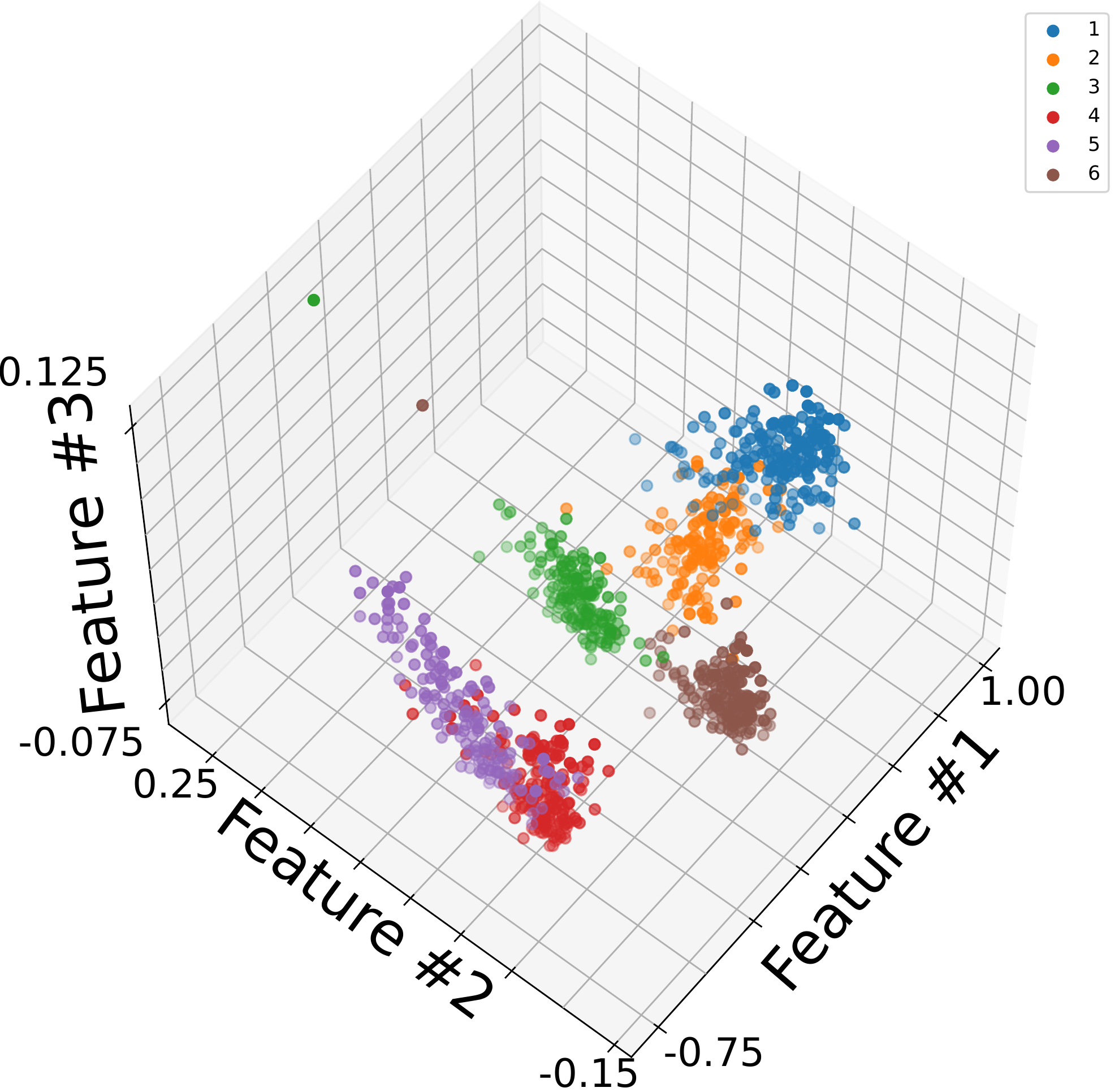}
  \end{center}
  \caption{Encodings of the vector fields obtained from the {\em Symbols} time series data set.}
  \label{fig:symbols_emb}
\end{figure}

We use the methods presented in the previous sections to obtain encodings of the vector fields generated from the {\em Symbols} time series data set. To train the Skip-gram model we use only the training set. In Figure~\ref{fig:symbols_emb} we show the encodings for the test set, where each dot represents a time series and colors represent classes. This visualization suggest that the method can be used in time series classification problem.
We may treat an encoding as a low dimensional vectorization of a time series and use it as an input to a classifier, e.g. SVM. This is a work in progress and we are going to publish more research results in the future.

\section{Conclusion and future work}
\label{sec:conc}
We introduced an efficient algorithm to compute hidden features of sampled vector fields. Our method utilizes modern machine learning techniques and provide new possibilities to study dynamical systems. By examples we show that the technique is able to find good characteristics of a given sampled vector field as well as collection of such fields.

We show applications of the method for a synthetic and experimental data sets. Our next step is to use the method to analyze bigger data sets, in particular: the magnetic field at the solar surface, time series data transformed into vector fields, velocity fields of a fluid flow measured by particle image velocimetry. Our main goal is to use the extracted features of vector fields as an input to other machine learning methods, e.g. SVM, decision trees, etc. We believe it will become a complementary description of a data set, useful in automatic analysis pipelines.

We would like to thank the anonymous reviewers for their comments and valuable suggestions.



\end{document}